\documentclass[10pt,twocolumn,letterpaper]{article}
\usepackage[pagenumbers]{paperstyle}

\usepackage{cuted}

\usepackage{currfile}

\usepackage{caption}
\usepackage{float}
\usepackage{placeins}
\usepackage{needspace}
\usepackage{microtype}

\AddToHook{build/column/before}{%
  \ifnum\value{page}=7\relax
    \raggedbottom
  \else
    \flushbottom
  \fi
}

 \usepackage{booktabs}
\usepackage{tabularx}
\usepackage{multirow}
\usepackage{colortbl}
\usepackage{array}
\usepackage{stfloats}
\definecolor{tbbest}{RGB}{244,194,194}
\definecolor{tbsecond}{RGB}{251,224,204}
\newcommand{\fst}[1]{\cellcolor{tbbest}\textbf{#1}}
\newcommand{\snd}[1]{\cellcolor{tbsecond}#1}
\newcommand{\oursrow}{}
\definecolor{suppbest}{RGB}{244,194,194}
\definecolor{suppsecond}{RGB}{251,224,204}
\newcommand{\bestnum}[1]{\cellcolor{suppbest}\textbf{#1}}
\newcommand{\secondnum}[1]{\cellcolor{suppsecond}#1}
\newcommand{\ranklegend}{Red and orange cells mark the best and second-best numerical results for each metric,
respectively.\xspace}
\newcommand{\notapplicable}{--}
\graphicspath{{figures/}}
\definecolor{linkblue}{rgb}{0.21,0.49,0.74}
\usepackage[pagebackref,breaklinks,colorlinks,allcolors=linkblue]{hyperref}

\title{RiskStage: Low-Cost Scene-Adaptive Generation of Safety-Critical Driving Scenarios with Semantic Feed-Forward Gaussians}
\author{{\normalsize
\begin{tabular}{@{}c@{}}
{\large An Xie,\quad Haotian Shi\textsuperscript{*},\quad Siying Li,\quad Yujia Zhao,\quad Jie Sun,\quad Jian Sun}\\[4pt]
College of Transportation Engineering, Tongji University\\
Key Laboratory of Road and Traffic Engineering, Ministry of Education\\[2pt]
{\small \textsuperscript{*}Corresponding author: \href{mailto:shihaotian95@tongji.edu.cn}{\texttt{shihaotian95@tongji.edu.cn}}}
\end{tabular}}}
\hypersetup{pdfauthor={An Xie, Haotian Shi, Siying Li, Yujia Zhao, Jie Sun, Jian Sun}}
\makeatletter
\patchcmd{\@maketitle}{\vspace*{24pt}}{\vspace*{6pt}}{}{\PackageError{arxiv}{Could not adjust title spacing}{}}
\makeatother

\begin{document}
\flushbottom
\maketitle
\setlength{\stripsep}{3pt plus 1pt minus 1pt}
\begin{strip}\centering
\includegraphics[width=0.99\textwidth]{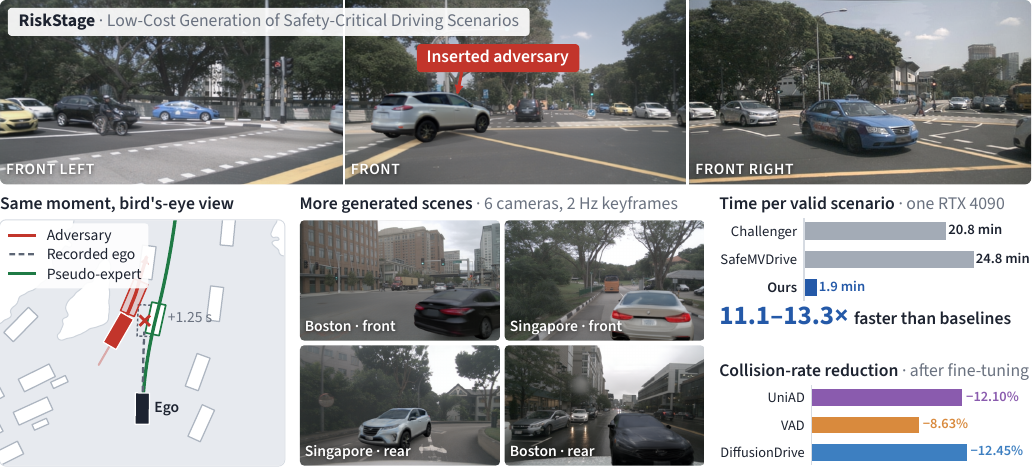}
\captionsetup{hypcap=false}
\captionof{figure}{\textbf{RiskStage} generates safety-critical driving scenario data through semantic reconstruction,
scene-adaptive interaction generation, and harmonized multi-view rendering, with pseudo-expert ego trajectories
for supervision.}
\label{fig:teaser}
\captionsetup{hypcap=true}
\end{strip}

\begin{abstract}
Safety-critical scenarios are essential for testing and training end-to-end (E2E) driving models, yet generating
them at scale with realistic appearance while keeping them solvable for the ego vehicle remains challenging. We
present RiskStage, a framework for low-cost generation of safety-critical 3D driving scenarios. The framework first
uses Semantic Feed-Forward Gaussians (SFG) to reconstruct a unified, editable 3D representation of geometry,
appearance, and semantics from multi-view images without scene-specific optimization. Building on this
representation, Scene-Adaptive Interaction Generation (SAIG) constructs challenging yet solvable interactions.
It initializes an adversarial vehicle based on road geometry and surrounding traffic and guides trajectory search by
evaluating ego avoidance responses under full-scene constraints. Coordinated Generation Scheduling (CGS)
coordinates computation across SFG-based scene reconstruction and multi-view observation synthesis for
SAIG-generated interactions to reduce computational cost. The resulting scenarios pair synchronized multi-view
observations with pseudo-expert ego trajectories, supporting both testing and training. Extensive
experiments on nuScenes demonstrate that RiskStage generates visually realistic scenarios with diverse
interactions while reducing GPU time per valid scenario by a factor of $11.1\text{--}13.3$ compared with
video-based baselines. Evaluations on three E2E driving models show that the generated scenarios expose model
limitations. Compared with controls fine-tuned on original-scene renderings, models fine-tuned with these data
achieve relative reductions of $8.6\%\text{--}12.5\%$ in open-loop collision rates on the adversarial test set.
Fine-tuning with these data can also improve closed-loop driving performance.
\end{abstract}
\section{Introduction}
\label{sec:intro}

End-to-end (E2E) autonomous driving models have advanced rapidly in recent years and become a major research
focus~\cite{hu2023uniad,jiang2023vad,liao2025diffusiondrive}. However, these models still exhibit safety and
reliability limitations in complex traffic scenarios~\cite{jia2024bench2drive}. One key reason is that existing
driving datasets predominantly capture routine traffic conditions, with limited coverage of safety-critical scenarios that
reveal model weaknesses and support targeted improvements~\cite{caesar2020nuscenes,ding2023survey}. Acquiring
such scenario data is therefore essential for testing and training E2E driving models.

However, acquiring safety-critical scenarios is not straightforward. Compared with costly and time-consuming
real-world data collection, simulation offers a practical route to acquiring such scenario data at low cost
and at scale~\cite{ge2025unraveling}. Diffusion-based methods can generate realistic and diverse
multi-view driving videos~\cite{gao2024magicdrive,gao2025magicdrivev2,challenger,safemvdrive}. However, these methods
are computationally expensive, and scene changes typically require generating videos from scratch, limiting
efficient reuse of previously generated content and making low-cost generation at scale difficult. They may also
exhibit inconsistencies across views and over long sequences. In contrast, methods based on 3D Gaussian Splatting
(3DGS) provide an explicit scene representation, allowing reconstructed geometry and appearance to be reused
through scene editing to create different driving scenarios~\cite{kerbl20233dgs,yan2024street,wang2026horizonforge}. Shared geometry
and object motion also support spatiotemporally consistent rendering. However, optimization-based
3D Gaussian methods require iterative optimization for each scene and therefore still incur substantial
reconstruction costs. More recently, feed-forward Gaussian methods have enabled direct reconstruction of scene
geometry and appearance from images without scene-specific optimization~\cite{recondrive,tian2025drivingforward,miao2025evolsplat}.
Their fast reconstruction and efficient rendering provide a technical foundation for generating safety-critical driving
scenarios at low cost and at scale.

Despite this promise, three challenges remain in using feed-forward Gaussians for low-cost safety-critical
driving scenario generation. First, scene editing requires
identifying drivable areas and traffic participants to determine where and what to modify. Existing feed-forward
driving scene reconstruction methods primarily recover geometry and appearance but lack the semantic information
needed to identify editing regions and objects~\cite{recondrive,tian2025drivingforward}. Meanwhile, Gaussian-based
semantic occupancy prediction provides spatial semantics but does not aim to recover renderable appearance,
limiting its direct use for rendering edited scenes~\cite{huang2024gaussianformer,jiang2025gausstr}.
A key challenge is therefore to unify geometry,
appearance, and semantics within a feed-forward representation for driving scene editing. Second, scene
reconstruction primarily reproduces routine traffic interactions in the source data, while constructing new
high-value adversarial interactions requires scene editing~\cite{ding2023survey,challenger}. The ego vehicle faces challenges not only
from the adversary but also from road geometry and surrounding traffic that restrict its avoidance
options~\cite{liu2025scenge}. Even plausible adversary motion may yield interactions that are
insufficiently challenging when the ego vehicle has ample room to maneuver, or unsolvable when no feasible
avoidance response exists~\cite{rempe2022strive,chen2024frea}. The second challenge
is thus to turn these scene constraints to the adversary's advantage when selecting its initial state and
trajectory, creating high-value interactions that are both challenging and solvable. Finally, generating
safety-critical scenarios with feed-forward Gaussians involves multiple interrelated stages, and fast
reconstruction does not directly guarantee efficient scenario generation. The third challenge is to control
the computational costs of multi-stage processing.

To address these challenges, we propose RiskStage (\cref{fig:teaser}), a framework for generating safety-critical 3D
driving scenarios at low cost and at scale. We first develop Semantic Feed-Forward Gaussians (SFG), a model that
reconstructs an editable 3D scene representation from multi-view images, unifying geometry, appearance, and
semantics. Building on this representation, we introduce Scene-Adaptive Interaction Generation (SAIG), which initializes an inserted
adversarial vehicle using road geometry and surrounding traffic and guides trajectory search by evaluating
ego avoidance responses under full-scene constraints to construct challenging yet solvable interactions.
To improve generation efficiency, we develop Coordinated Generation Scheduling (CGS), which coordinates
computation across scenario generation stages through strategies such as shared visual encoding and adaptive
image harmonization. Finally, we construct pseudo-expert ego trajectories as supervision for the generated
scenarios and pair them with synchronized multi-view observations for testing and training E2E driving models.

Our main contributions are summarized below.
\begin{itemize}
\item We propose RiskStage, a low-cost framework for safety-critical 3D driving scenario generation that provides high-value
scenario data for testing and training E2E driving models.
\item We develop SFG, a semantic feed-forward Gaussian model that reconstructs a unified, editable 3D
representation of geometry, appearance, and semantics from multi-view images.
\item We design SAIG to couple scene-adaptive initialization with full-scene solvability guidance during
trajectory search, constructing challenging yet solvable interactions.
\item We introduce CGS, which uses shared visual encoding and adaptive image harmonization to coordinate
computation across stages and improve generation efficiency.
\end{itemize}
\section{Related Work}
\label{sec:related}
\paragraph{Driving Scene Reconstruction and Simulation.}
Radiance-field and Gaussian-based methods reconstruct static backgrounds and moving vehicles from recorded
drives, enabling novel-view synthesis and vehicle editing~\cite{yang2023unisim,yan2024street,chen2025omnire}.
Dynamic Gaussian representations capture temporal scene changes~\cite{chen2023pvg}. Reconstructed scenes also
provide a basis for photorealistic closed-loop simulation~\cite{zhou2024hugsim}. However, scene-specific
optimization makes large-scale reconstruction costly. Feed-forward reconstruction enables direct prediction
of scene geometry, camera parameters~\cite{wang2025vggt}, and renderable Gaussian
representations~\cite{charatan2024pixelsplat,chen2024mvsplat} from images. Driving-oriented feed-forward methods
support reconstruction from flexible surround-view inputs~\cite{tian2025drivingforward} and sparse ego-centric
views~\cite{wei2025omniscene}. Multi-view feature aggregation and explicit motion modeling further enable
dynamic scene reconstruction~\cite{recondrive}. We build on renderable feed-forward Gaussian representations
for driving scene editing and safety-critical scenario generation.

\paragraph{Semantic Gaussian Representations.}
Semantic Gaussian methods jointly model geometry, appearance, and semantics~\cite{zhou2024hugs,he2025jointsemantic},
or distill 2D semantic and language features into Gaussians for querying and
editing~\cite{zhou2024feature3dgs,qin2024langsplat}. These approaches rely on scene-specific optimization. Generalizable methods
predict semantic Gaussians from uncalibrated image pairs using 2D supervision, without scene-specific
optimization~\cite{gsemsplat}. In autonomous driving, Gaussian-based semantic occupancy methods predict spatial
occupancy and semantic categories from multi-view images~\cite{huang2024gaussianformer,jiang2025gausstr}. These
occupancy prediction methods focus on spatial occupancy and semantics rather than reconstructing appearance for
image rendering. SFG instead binds semantic predictions from shared reconstruction features to renderable Gaussians
for identifying editing regions and objects.

\paragraph{Driving Scenario Generation.}
Trajectory-level methods construct safety-critical interactions using kinematic-gradient
optimization~\cite{hanselmann2022king}, learned traffic priors~\cite{rempe2022strive}, and adversarial policy
learning~\cite{chen2024frea}. Guided diffusion supports rule-controlled traffic simulation~\cite{zhong2023ctg}
and reactive closed-loop adversaries~\cite{chang2024safesim}. To retain solvable scenarios, some methods search
for safe ego trajectories after adversarial generation~\cite{rempe2022strive}, while others incorporate a
learned ego feasible region into policy optimization~\cite{chen2024frea}. Background vehicle trajectories have
also been optimized to restrict the ego vehicle's avoidance options and create occlusions~\cite{liu2025scenge}.
At the sensor level, adversarial simulation updates LiDAR observations to match perturbed vehicle
trajectories~\cite{wang2021advsim}. For camera-based E2E models, recent methods combine trajectory generation
with diffusion-based multi-view video synthesis~\cite{challenger,safemvdrive}. One approach penalizes collisions
with the recorded ego trajectory~\cite{challenger}, whereas another first constructs a collision scenario and
then optimizes an ego avoidance trajectory~\cite{safemvdrive}. However, video synthesis incurs substantial
computational cost~\cite{challenger}. Our framework uses scene-adaptive vehicle initialization and evaluates
parameterized ego avoidance responses under full-scene constraints during trajectory search, producing
multi-view scenario data at low cost.
\section{Method}
\label{sec:method}
\begin{figure*}[t]\centering
\includegraphics[width=0.92\textwidth]{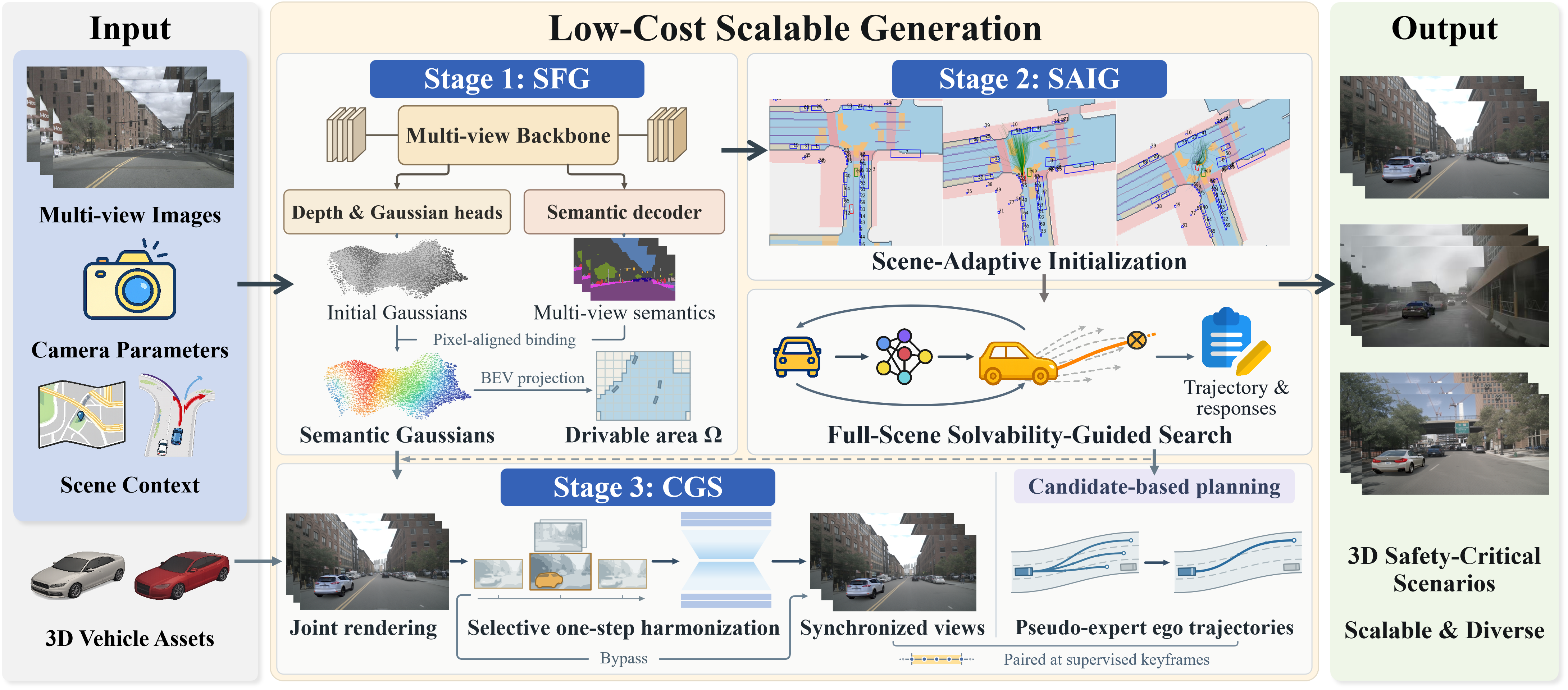}
\caption{\textbf{RiskStage overview.} SFG provides editable scenes for SAIG to construct adversarial interactions,
while CGS coordinates computation across generation stages to improve efficiency.}
\label{fig:overview}
\end{figure*}

This section presents the RiskStage framework (\cref{fig:overview}). We first formulate the scenario generation
task and then describe the designs of SFG, SAIG, and CGS.

\subsection{Problem Formulation}
\label{sec:formulation}

We take as input a recorded drive $\mathcal{D}=\{(I_k,T_k,\mathcal{B}_k)\}_{k=1}^{K}$ and a library of 3D vehicle
assets. Here, $K$ is the number of keyframes, and $I_k$, $T_k$, and $\mathcal{B}_k$ denote the calibrated
multi-view images, ego pose, and set of 3D bounding boxes of surrounding traffic at keyframe $k$, respectively.

The generated scenario is
\begin{equation}
\mathcal{S}=\big(\tau,\;\{\hat I_k\}_{k=1}^{K},\;\{y_k\}_{k\in\mathcal{W}}\big),
\label{eq:scenario_output}
\end{equation}
where $\tau$ is the adversarial vehicle trajectory, $\hat I_k$ the edited multi-view observations, and $y_k$ a
pseudo-expert future ego trajectory starting at keyframe $k$. The supervision window
$\mathcal{W}\subseteq\{1,\ldots,K\}$ specifies the keyframes with ego-trajectory supervision. Our goal is to
generate, at low computational cost, safety-critical scenarios that are challenging yet solvable for the ego
vehicle, providing data for testing and training E2E driving models.

\subsection{Semantic Feed-Forward Gaussians}
\label{sec:semantic}

\paragraph{Unified scene representation.}
SFG reconstructs a unified, pixel-aligned Gaussian representation of geometry, appearance, and semantics from
multi-view images for driving scene editing. We initialize the reconstruction branch from a pretrained
feed-forward model~\cite{recondrive}. The representation is
\begin{equation}
\mathcal{G}=\{g_i\}_{i=1}^{N_G},\quad
g_i=(\boldsymbol{\mu}_i,\boldsymbol{\Sigma}_i,\alpha_i,\mathbf{c}_i,\mathbf{p}_i),
\label{eq:semantic_gaussians}
\end{equation}
where $N_G$ is the number of Gaussians. Each Gaussian comprises a
center $\boldsymbol{\mu}_i$, covariance $\boldsymbol{\Sigma}_i$, opacity $\alpha_i$, appearance coefficients
$\mathbf{c}_i$, and a semantic distribution $\mathbf{p}_i$ over $C$ classes.

\paragraph{Semantic prediction and joint fine-tuning.}
The semantic decoder~\cite{ranftl2021dpt} reuses multi-view features from the reconstruction backbone and
combines them with input RGB and ego-frame height derived from predicted depth. It predicts pixelwise class
probabilities $\mathbf{p}_i$, which are bound to the corresponding Gaussians.

We jointly fine-tune the shared backbone, depth and Gaussian reconstruction branches, and semantic decoder using
\begin{equation}
\mathcal{L}
=\mathcal{L}_{\mathrm{rec}}+\mathcal{L}_{\mathrm{src}}
+\lambda_{\mathrm{aux}}\mathcal{L}_{\mathrm{aux}}
+\lambda_{\mathrm{rend}}\mathcal{L}_{\mathrm{rend}},
\label{eq:joint_loss}
\end{equation}
where $\mathcal{L}_{\mathrm{rec}}$ retains the original reconstruction objective. The source-view loss
$\mathcal{L}_{\mathrm{src}}$ is computed only at pixels with projected semantic annotations, with coarse labels
supervising the corresponding class unions. The auxiliary loss $\mathcal{L}_{\mathrm{aux}}$ uses confident
predictions from a fixed segmentation teacher~\cite{borse2021inverseform} only where annotations are absent.
The rendered-view loss $\mathcal{L}_{\mathrm{rend}}$ supervises normalized semantic renderings using target-view
annotations and constrains Gaussian geometry through differentiable rendering. The teacher model is not
required at inference.

\paragraph{Drivable-area extraction.}
At inference, geometry and semantics are first computed to construct the scene constraints for SAIG. We project
road-labeled Gaussian centers onto a bird's-eye-view raster in a common coordinate frame and supplement road
coverage with recorded vehicle footprints. The resulting drivable area $\Omega$ guides vehicle initialization
and ego-response evaluation in SAIG.

\subsection{Scene-Adaptive Interaction Generation}
\label{sec:pressure}
\begin{figure*}[t]\centering
\includegraphics[width=0.96\textwidth]{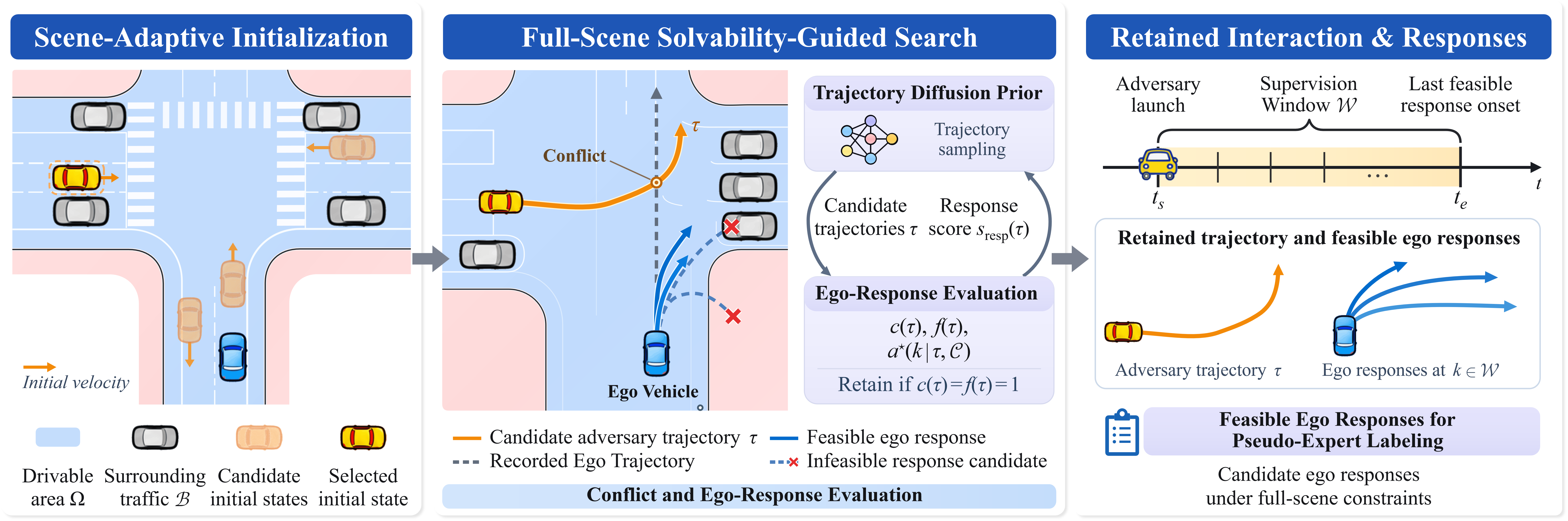}
\caption{SAIG comprises scene-adaptive initialization, full-scene solvability-guided trajectory search,
and interaction retention with ego-response output.}
\label{fig:method_detail}
\end{figure*}

Road geometry and surrounding traffic restrict the ego vehicle's avoidance options. SAIG uses these scene
conditions in three steps, namely scene-adaptive initialization, full-scene solvability-guided trajectory search, and
interaction retention with ego-response output (\cref{fig:method_detail}).

\paragraph{Scene-adaptive initialization.}
We represent a candidate initial state as $\xi=(t_s,\mathbf{x},\psi,v)$, specifying launch time, two-dimensional
position, heading, and speed. Positions are sampled within $\Omega$, headings follow local road geometry and
traffic direction, and speeds are set based on nearby vehicle motion. After road-coverage and traffic-clearance
checks, we score feasible candidates by
\begin{equation}
s_{\mathrm{init}}(\xi)=\mathbf{w}^{\top}\boldsymbol{\phi}(\xi;\Omega,\mathcal{B}).
\label{eq:initialization_score}
\end{equation}
Here, $\mathcal{B}$ denotes surrounding traffic following recorded trajectories. The feature vector
$\boldsymbol{\phi}$ includes road coverage, directional compatibility, and speed consistency, with preset
weights $\mathbf{w}$. We prioritize high-scoring states while removing near-duplicates and retaining candidates
across travel directions and launch times.

\paragraph{Full-scene solvability-guided trajectory search.}
Starting from the selected initial states, a pretrained trajectory diffusion prior proposes candidate trajectories
through iterative sampling and refinement~\cite{challenger}. For each $\tau$, we evaluate parameterized ego avoidance responses combining
longitudinal speed changes and lateral displacement relative to recorded ego motion. Under
$\mathcal{C}=(\Omega,\mathcal{B})$, feasible responses must remain within $\Omega$, avoid collisions with the
adversarial vehicle and surrounding traffic, and satisfy vehicle motion constraints.

Let $a^\star(k\mid\tau,\mathcal{C})$ denote the minimum additional acceleration magnitude among feasible responses
starting at keyframe $k$, taking $\infty$ if none is found. The binary indicator $c(\tau)$ denotes a collision
with the ego vehicle following its recorded motion. We set $f(\tau)=1$ when $c(\tau)=1$ and a feasible response
is found at a sampled onset at or after adversarial vehicle launch and before the first collision; otherwise,
$f(\tau)=0$. Avoidability is assessed within the evaluated response family.

At each replanning step, we rank candidates by
$J(\tau)=J_{\mathrm{base}}(\tau)+\lambda_{\mathrm{resp}}s_{\mathrm{resp}}(\tau)$ and use lower-scoring candidates
to guide subsequent search. The response term is
\begin{equation}
s_{\mathrm{resp}}(\tau)=(1-c(\tau))(1+\widetilde d(\tau))+\gamma c(\tau)(1-f(\tau)).
\label{eq:response_score}
\end{equation}
Here, $J_{\mathrm{base}}$ penalizes road violations and collisions with surrounding traffic, and
$\lambda_{\mathrm{resp}}$ weights the response term. The term $\widetilde d(\tau)\in[0,1]$ is the normalized
minimum distance to the recorded ego trajectory, while $\gamma$ penalizes conflicts for which no feasible
response is found.

\paragraph{Interaction retention and ego-response output.}
We retain interactions whose adversarial trajectories satisfy road and motion constraints, avoid collisions
with surrounding traffic, and meet $c(\tau)=f(\tau)=1$. The supervision window $\mathcal{W}$ includes keyframes
from $t_s$ to the last feasible response onset $t_e$. Retained feasible ego responses provide candidates for
subsequent pseudo-expert ego trajectory generation.

\subsection{Coordinated Generation Scheduling}
\label{sec:render}

CGS reduces generation cost through shared visual encoding and adaptive image harmonization. For multi-view
inputs $\mathcal{I}$, CGS retains the backbone features used for semantic prediction and reuses them in subsequent
Gaussian reconstruction, giving $\mathbf{F}_{\mathrm{sem}}=\mathbf{F}_{\mathrm{rec}}=E(\mathcal{I})$, where $E$
denotes the shared backbone encoder. For each retained trajectory $\tau$, we jointly render the reconstructed
scene and the adversarial vehicle's Gaussian asset~\cite{du2025realcar,cao2026assetharvester} to obtain
$\tilde I_{k,j}$, the image from camera $j$ at keyframe $k$. CGS adaptively selects frames for harmonization
according to the visible effect of the edit,
\begin{equation}
\mathcal{Q}_{\tau}=\operatorname{Expand}_{\mathrm{time}}
\bigl\{(k,j)\mid A_{k,j}(\tau)\ge\delta\bigr\},
\label{eq:harmonization_selection}
\end{equation}
where $A_{k,j}(\tau)$ counts pixels changed by the edit relative to the corresponding rendering without the
adversarial vehicle. The threshold $\delta$ controls frame selection, and $\operatorname{Expand}_{\mathrm{time}}$
adds neighboring frames needed for temporal context. The final observations are
\begin{equation}
\hat I_{k,j}=
\begin{cases}
\mathcal{H}(\tilde I_{k,j};\mathcal{U}_{k,j}), & (k,j)\in\mathcal{Q}_{\tau},\\
\tilde I_{k,j}, & \text{otherwise}.
\end{cases}
\label{eq:harmonized_observations}
\end{equation}
Here, $\mathcal{H}$ is a pretrained single-step diffusion harmonizer~\cite{zhang2026diffusionharmonizer}, and
$\mathcal{U}_{k,j}$ provides temporal context from previously harmonized frames of the same camera. Outputs
from all cameras form the synchronized multi-view observations $\hat I_k$.

We use candidate-based planning~\cite{dauner2023pdm} to jointly select pseudo-expert ego trajectories $y_k$
across keyframes in $\mathcal{W}$ from recorded ego motion, car-following trajectories~\cite{treiber2000idm}, and feasible avoidance
responses retained by SAIG. We assess candidates for collision avoidance, road compliance, and motion
feasibility and pair the selected trajectories with the corresponding observations $\hat I_k$ to provide
ego-trajectory annotations.
\FloatBarrier
\section{Experiments}
\label{sec:experiments}
\begin{figure*}[t]\centering
\begin{minipage}[t]{0.485\textwidth}\centering
\includegraphics[width=0.96\linewidth]{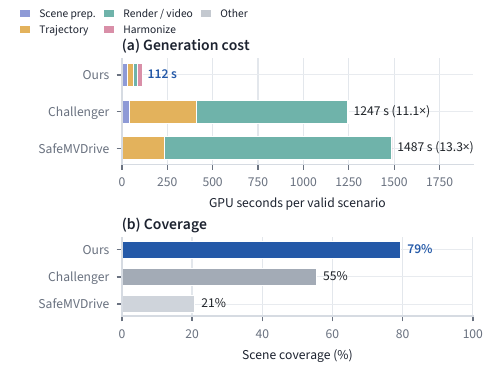}
\caption{\textbf{Generation cost and coverage.} (a) Time per valid scenario. (b) Scene coverage.}
\label{fig:cost}
\end{minipage}\hfill
\begin{minipage}[t]{0.485\textwidth}\centering
\includegraphics[width=0.96\linewidth]{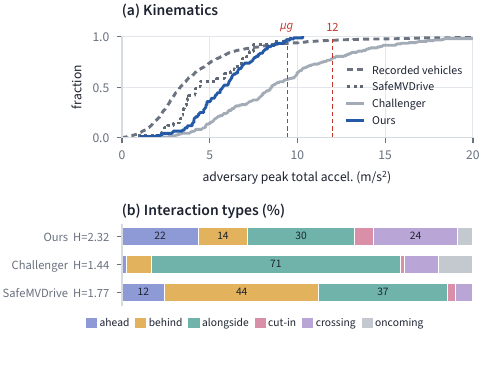}
\caption{\textbf{Motion and diversity.} (a) Peak acceleration ($\mu g=9.43$\,m/s$^2$). (b) Interaction types.}
\label{fig:adversary}
\end{minipage}
\end{figure*}

We conduct experiments to evaluate RiskStage. The experimental setup is presented first (\cref{sec:setup}).
We then assess generation quality and cost (\cref{sec:generation}). Building on this assessment, we evaluate
the scenarios' practical utility through testing and training E2E driving models (\cref{sec:value}). Finally,
we examine individual designs through component analysis and ablations (\cref{sec:components}).

\subsection{Experimental Setup}
\label{sec:setup}

\paragraph{Data.}
We conduct experiments on nuScenes~\cite{caesar2020nuscenes}, following the official split of 700 training and
150 validation scenes and using six-camera observations at 2\,Hz.

\paragraph{Comparisons.}
We compare generation quality, cost, and coverage across methods~\cite{challenger,safemvdrive} on the nuScenes
validation split. We further assess scene representation quality against task-specific RGB
reconstruction~\cite{yan2024street,chen2025omnire,tian2025drivingforward} and semantic segmentation baselines.
We report PSNR, SSIM~\cite{wang2004ssim}, and LPIPS~\cite{zhang2018lpips} for reconstruction, and
FID~\cite{heusel2017fid} and FVD~\cite{unterthiner2018fvd} for generated observations.
Runtime is measured on a single RTX\,4090.

\paragraph{E2E model evaluation.}
We evaluate UniAD~\cite{hu2023uniad}, VAD~\cite{jiang2023vad}, and DiffusionDrive~\cite{liao2025diffusiondrive}
using open-loop collision rate and closed-loop HD-Score in HUGSIM~\cite{zhou2024hugsim}.

\subsection{Generation Quality and Cost}
\label{sec:generation}
\label{sec:recon}
\paragraph{Generation cost and coverage.}
RiskStage generates a valid safety-critical scenario in 112 seconds on average, achieving
$11.1$--$13.3\times$ speedups over the baselines (\cref{fig:cost}a). It produces at least one valid
scenario for 119 of the 150 validation scenes, yielding a scene coverage (Cov.) of 79.3\%, 24.0 percentage points
above the highest baseline coverage (\cref{fig:cost}b).

\begin{table}[!htb]\centering\footnotesize\setlength{\tabcolsep}{1.05pt}
\caption{\textbf{Scene reconstruction quality and per-scene runtime.}}
\label{tab:recon}
\begin{tabularx}{\columnwidth}{>{\raggedright\arraybackslash}Xccccccc}\toprule
 & \multicolumn{3}{c}{Recon.} & \multicolumn{3}{c}{Novel} & \\
\cmidrule(lr){2-4}\cmidrule(lr){5-7}
Method & PSNR$\uparrow$ & SSIM$\uparrow$ & LPIPS$\downarrow$ & PSNR$\uparrow$ & SSIM$\uparrow$ & LPIPS$\downarrow$ & Time$\downarrow$ \\\midrule
Street Gaussians & 29.16 & 0.882 & 0.166 & 22.96 & \snd{0.696} & 0.295 & 20\,m \\
OmniRe & 29.42 & 0.885 & 0.158 & 23.01 & 0.689 & \fst{0.276} & 21\,m \\\midrule
DrivingForward & 22.83 & 0.765 & 0.256 & 21.88 & 0.687 & 0.298 & \fst{6\,s} \\
ReconDrive & \snd{33.26} & \snd{0.957} & \snd{0.064} & \fst{23.15} & \fst{0.697} & \snd{0.292} & \snd{21\,s} \\
\oursrow\textbf{Ours} & \fst{34.05} & \fst{0.960} & \fst{0.059} & \snd{23.12} & \snd{0.696} & 0.293 & 22\,s \\\bottomrule
\end{tabularx}
\end{table}

\begin{table}[!htb]\centering\small\setlength{\tabcolsep}{3.2pt}
\caption{\textbf{Generation quality, cost, and coverage.} Time is measured in seconds per valid scenario.
Red/orange: best/second-best.}
\label{tab:generation}
\begin{tabularx}{\columnwidth}{>{\raggedright\arraybackslash}Xcccccc}\toprule
Method & Time$\downarrow$ & Cov.$\uparrow$ & FID$\downarrow$ & FVD$\downarrow$ & $>\!\mu g\downarrow$ & $H\uparrow$ \\\midrule
Challenger & \snd{1247} & \snd{55\%} & 25.94 & \snd{80.86} & 41.9\% & 1.44 \\
SafeMVDrive & 1487 & 21\% & \snd{23.31} & 131.63 & \fst{2.4\%} & \snd{1.77} \\
\oursrow\textbf{Ours} & \fst{112} & \fst{79\%} & \fst{6.78} & \fst{16.62} & \snd{3.3\%} & \fst{2.32} \\\bottomrule
\end{tabularx}
\end{table}

\begin{figure*}[t]\centering
\includegraphics[width=0.96\textwidth]{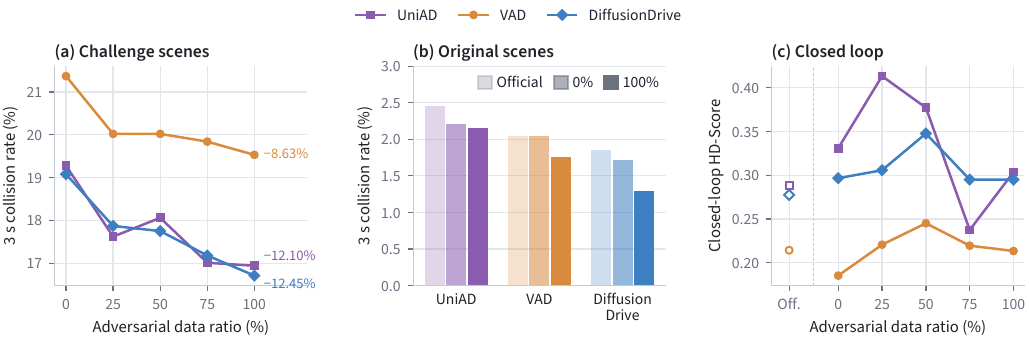}
\caption{\textbf{Effect of adversarial-data fine-tuning.} (a,b) Adversarial and original test sets, respectively;
(c) 10 HUGSIM scenarios. Ratios denote fine-tuning mixtures; test sets are fixed. The 0\% control uses
original-scene renderings; Official/Off.\ denotes pretrained weights. Annotations in (a) show relative
collision-rate reductions from 0\% to 100\%.}
\label{fig:training}
\end{figure*}

\begin{figure*}[t]\centering
\includegraphics[width=0.96\textwidth]{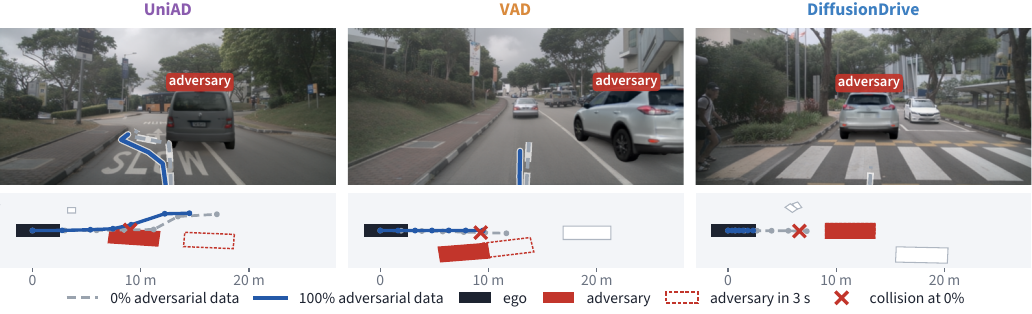}
\caption{\textbf{Open-loop plans after fine-tuning.} Top and bottom show front-camera and bird's-eye views of
3\,s plans. The 0\% control uses original-scene renderings; 100\% uses the adversarial version.
Training iterations are matched within each model.}
\label{fig:planning}
\end{figure*}

\begin{figure*}[t]\centering
\includegraphics[width=0.96\textwidth]{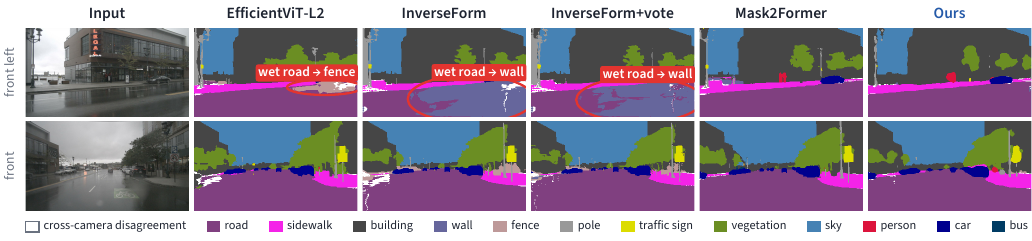}
\caption{\textbf{Cross-camera semantic predictions.} SFG and baselines are compared on two neighboring views
at the same time. White pixels mark inconsistent predictions for corresponding surfaces across cameras.}
\label{fig:semantic}
\end{figure*}

\paragraph{Visual quality.}
We first evaluate the reconstruction quality of the 3D representation used for scene editing, following the
input-view (Recon.) and novel-view (Novel) protocol in~\cite{recondrive}. Joint fine-tuning increases input-view
PSNR from 33.26 to 34.05\,dB, while novel-view PSNR remains close to the frozen baseline
(23.12 versus 23.15\,dB). Reconstruction takes
22 seconds per scene, about $55\times$ faster than scene-specific optimization (\cref{tab:recon}). We then
evaluate the visual quality of the final observations after scene editing. These observations achieve lower
FID and FVD than both baselines, at 6.78 and 16.62, respectively (\cref{tab:generation}).

\paragraph{Motion and interaction diversity.}
We adopt an acceleration reference of 9.43\,m/s$^2$ based on published braking-test data~\cite{adac2024braking}.
Peak acceleration exceeds this value for 3.3\% of our adversarial vehicles, compared with 2.4\% and 41.9\%
for baselines (\cref{fig:adversary}a). Entropy over the six interaction types reaches $H=2.32$ bits versus at most
1.77 for baselines, indicating a more balanced distribution (\cref{fig:adversary}b).

Overall, RiskStage enables low-cost generation of safety-critical driving scenarios with good visual quality
and a balanced distribution of interaction types.

\subsection{Testing and Training Utility}
\label{sec:value}

{\looseness=-1
To assess the utility of scenarios generated by RiskStage for testing and training E2E driving models, we
construct training and test sets for original scenarios and their adversarial variants following the official
nuScenes split. The adversarial training and test sets contain 549 and 119 adversarial scenarios,
respectively. We use original-scene renderings for the remaining scenarios to match the official split sizes.
We evaluate training utility across data mixtures by fine-tuning models with adversarial-data proportions of
0\%, 25\%, 50\%, 75\%, and 100\%. The 0\% control uses only original-scene
renderings to distinguish the contribution of adversarial data from the effects of fine-tuning on re-rendered
observations alone.
\par}

\paragraph{Open-loop evaluation.}
Compared with the original scenarios, our generated safety-critical scenarios more effectively expose limitations
of E2E driving models. Using the same 0\% control models, we observe 3-second collision rates of 19.1--21.4\%
on the adversarial test set, compared with 1.7--2.2\% on the corresponding original test set (\cref{fig:training}a,b).

\paragraph{Model fine-tuning.}
Compared with the 0\% controls, fine-tuning with adversarial data can reduce collision rates for all three E2E
models on the adversarial test set (\cref{fig:training}a). At 100\%, UniAD, VAD, and DiffusionDrive achieve
relative collision-rate reductions of 12.10\%, 8.63\%, and 12.45\%, respectively. On the original test set,
all three models can also achieve lower collision rates than their respective 0\% controls
(\cref{fig:training}b), indicating that the generated adversarial data can also improve open-loop planning
safety in the original driving scenarios. \Cref{fig:planning} illustrates the resulting changes in open-loop
plans, including lateral avoidance by UniAD and longitudinal adjustments by VAD and DiffusionDrive.

\paragraph{Closed-loop evaluation.}
HUGSIM evaluation further shows that fine-tuning with our adversarial data can improve closed-loop driving performance
across all three E2E models (\cref{fig:training}c). For example, UniAD's HD-Score increases from 0.331 at 0\%
to 0.413 at 25\%.

{\widowpenalty=10000
\paragraph{Effect of adversarial-data proportion.}
Among the tested mixtures, all three models achieve their lowest \mbox{open-loop} collision rates on the adversarial
test set at 100\%, whereas their highest closed-loop HD-Scores occur at 25\% or 50\% (\cref{fig:training}a,c).
These results indicate that a higher adversarial-data proportion does not necessarily yield better
\mbox{closed-loop performance}.
\par}

\begingroup
\setlength{\intextsep}{8pt plus 2pt minus 2pt}
\subsection{Component Analysis and Ablation Studies}
\label{sec:components}
\label{sec:semantic_exp}
\label{sec:ablation}

We select 50 scenes from the nuScenes validation split to evaluate SFG and SAIG. Each SAIG configuration is
run with three random seeds per scene.

\begin{table}[!htb]\centering\small\setlength{\tabcolsep}{2.9pt}
\caption{\textbf{Semantic segmentation and drivable-area extraction.} Time is measured in ms per 12-view window.}
\label{tab:semantic}
\begin{tabularx}{\columnwidth}{>{\raggedright\arraybackslash}Xcccccc}\toprule
Source & Road & Veh. & Acc. & Cons. & Driv. & Time \\\midrule
EfficientViT-L2~\cite{cai2023efficientvit} & 0.895 & 0.711 & 0.859 & 0.915 & 0.653 & 278.4 \\
InverseForm     & \snd{0.905} & 0.877 & \snd{0.893} & 0.935 & \snd{0.666} & \snd{193.0} \\
InverseForm+vote & 0.903 & 0.864 & 0.890 & \fst{0.977} & -- & 195.2 \\
Mask2Former~\cite{cheng2022mask2former} & 0.888 & \snd{0.904} & \snd{0.893} & 0.932 & 0.654 & 883.4 \\
\oursrow\textbf{Ours} & \fst{0.954} & \fst{0.937} & \fst{0.945} & \snd{0.955} & \fst{0.674} & \fst{62.3} \\\bottomrule
\end{tabularx}
\end{table}

\begin{table}[!htb]\centering\footnotesize\setlength{\tabcolsep}{4pt}
\caption{\textbf{SAIG ablation.} Conflicts are adversary-compliant.
No response (\%) is per conflict; Valid (\%) is per 150 attempts.}
\label{tab:ablation}
\begin{tabularx}{\columnwidth}{>{\raggedright\arraybackslash}Xccc}\toprule
\multicolumn{4}{l}{\textit{(a) Initialization (full-scene guidance fixed)}} \\\midrule
Initialization & Conflicts & $H\uparrow$ & Valid (\%)$\uparrow$ \\\midrule
Non-adaptive legal & 66 & 2.01 & 42.7 \\
Pool-random selection & 82 & \snd{2.02} & \snd{52.7} \\
\oursrow\textbf{Adaptive selection} & 105 & \fst{2.26} & \fst{69.3} \\\midrule
\multicolumn{4}{l}{\textit{(b) Guidance (adaptive initialization fixed)}} \\\midrule
Guidance scope & Conflicts & No response (\%)$\downarrow$ & Valid (\%)$\uparrow$ \\\midrule
Adversary only & 107 & 15.0 & 60.7 \\
\oursrow\textbf{Full scene} & 105 & \fst{1.0} & \fst{69.3} \\\bottomrule
\end{tabularx}
\end{table}

\paragraph{Semantic representation.}
\Cref{tab:semantic} reports road and vehicle IoU (Road/Veh.), labeled-point accuracy (Acc.), and drivable-area
IoU (Driv.).
Compared with standalone 2D segmentation baselines~\cite{borse2021inverseform,cai2023efficientvit,cheng2022mask2former},
SFG improves semantic accuracy and cross-camera consistency at lower computational cost. Road and vehicle IoU
exceed the best baseline results by 4.9 and 3.3 percentage points, respectively, and semantic prediction is
$3.1$--$14.2\times$ faster (\cref{tab:semantic}). Cross-camera agreement
(Cons.) rises from 93.5\% to 95.5\%. \Cref{fig:semantic} qualitatively shows more accurate and consistent road
predictions.

\paragraph{SAIG ablation.}
With full-scene guidance fixed, adaptive initialization increases the conflict count from 82 to 105 and
interaction-type entropy from 2.02 to 2.26 relative to \mbox{random} selection from the same candidate pool
(\cref{tab:ablation}a). With adaptive initialization fixed, full-scene solvability guidance reduces the
proportion of conflicts with no feasible ego response found from 15.0\% to 1.0\% relative to adversary-only
guidance, with comparable conflict counts (\cref{tab:ablation}b).

\paragraph{CGS ablation.}
CGS reduces generation time per valid scenario from 205.2 to 122.6 seconds ($1.67\times$), with lower FID and FVD
(\cref{tab:cgs_ablation}).

\begin{table}[H]\centering\small\setlength{\tabcolsep}{6pt}
\caption{\textbf{CGS ablation.} Time is measured in seconds per valid scenario.
FID/FVD use matched six-camera frames.}
\label{tab:cgs_ablation}
\begin{tabularx}{\columnwidth}{l*{3}{>{\centering\arraybackslash}X}}\toprule
Setting & Time$\downarrow$ & FID$\downarrow$ & FVD$\downarrow$ \\\midrule
w/o CGS & 205.2 & 15.89 & 54.72 \\
\oursrow\textbf{CGS} & \fst{122.6} & \fst{10.65} & \fst{37.71} \\\bottomrule
\end{tabularx}
\end{table}
\endgroup
\FloatBarrier
\section{Conclusion}
\label{sec:conclusion}
We presented RiskStage, a framework for low-cost generation of safety-critical 3D driving scenarios.
SFG unifies geometry, appearance, and semantics into an editable scene representation. Building on this
representation, SAIG uses scene constraints to construct challenging yet solvable interactions. CGS coordinates
computation across generation stages to improve efficiency. Experiments show that RiskStage achieves an
$11.1$--$13.3\times$ speedup over the baselines in generating valid scenarios. The generated scenarios are
visually realistic, contain diverse interactions, and support the testing and training of E2E driving models.

Despite these results, SAIG assesses solvability within a parameterized ego-response family, while surrounding
traffic follows recorded trajectories. Future work will explore online closed-loop interaction generation
with reactive surrounding traffic.

\Needspace{6\baselineskip}
\begin{samepage}
\section*{LLM Usage Statement}
OpenAI Codex and Claude Code assisted with coding, data analysis, and manuscript preparation.
The authors verified all results and claims and are responsible for the work.
\end{samepage}

\clearpage
 
{\small
\fontsize{9}{10.6}\selectfont

}

\clearpage
\maketitlesupplementary
\appendix
\setcounter{figure}{0}
\setcounter{table}{0}
\setcounter{equation}{0}
\renewcommand{\thesection}{\Alph{section}}
\renewcommand{\thesubsection}{\thesection.\arabic{subsection}}
\renewcommand{\thetable}{S\arabic{table}}
\renewcommand{\thefigure}{S\arabic{figure}}
\renewcommand{\theequation}{S\arabic{equation}}
\renewcommand{\theHequation}{supp.\arabic{equation}}
\captionsetup[table]{skip=4pt}
\setcounter{topnumber}{3}
\setcounter{totalnumber}{4}
\setcounter{dbltopnumber}{3}
\renewcommand{\topfraction}{0.85}
\renewcommand{\dbltopfraction}{0.85}
\renewcommand{\textfraction}{0.12}
\renewcommand{\floatpagefraction}{0.65}
\renewcommand{\dblfloatpagefraction}{0.65}
\setlength{\floatsep}{9pt plus 2pt minus 2pt}
\setlength{\textfloatsep}{12pt plus 3pt minus 2pt}
\setlength{\intextsep}{9pt plus 2pt minus 2pt}
\setlength{\dblfloatsep}{9pt plus 6pt minus 2pt}
\setlength{\dbltextfloatsep}{12pt plus 6pt minus 2pt}
\AddToHook{cmd/section/before}{\Needspace*{4\baselineskip}}
\AddToHook{cmd/subsection/before}{\Needspace*{4\baselineskip}}

\looseness=-1
This supplementary material provides implementation details and more extensive experimental evaluations of RiskStage.
\Cref{supp:sec:sfg,supp:sec:saig,supp:sec:cgs} describe SFG, SAIG, and CGS, respectively.
\Cref{supp:sec:generation_protocol} details the evaluation settings and how the metrics are computed for scenario generation,
and \cref{supp:sec:e2e} presents pseudo-expert ego trajectory construction, E2E driving model fine-tuning,
and evaluation results. We will release the code upon publication.

\section{Semantic Feed-Forward Gaussians}
\label{supp:sec:sfg}

This section supplements the SFG architecture and joint-training formulation in the main paper.
We describe semantic binding and supervision, examine the effect of joint training, and evaluate
reconstruction, semantic segmentation, and cross-dataset generalization.

\subsection{Architecture and Semantic Binding}
\label{supp:sec:sfg_architecture}

\begin{figure*}[!t]
\centering
\includegraphics[width=\textwidth]{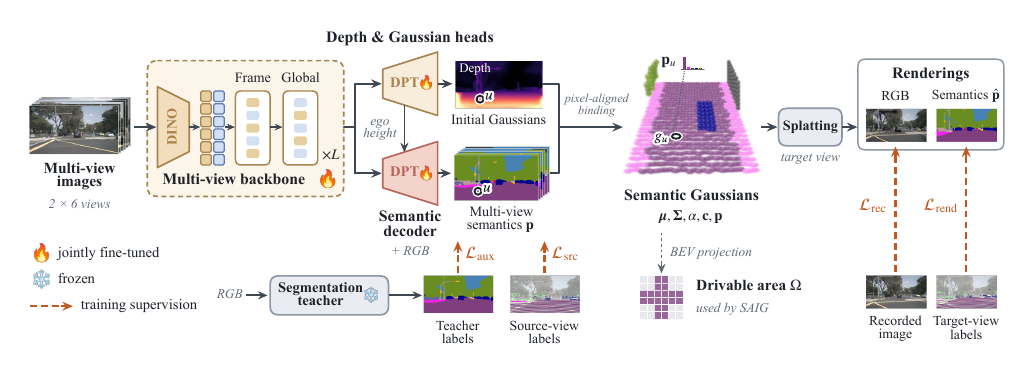}
\caption{\textbf{SFG architecture and joint training.} Gaussian reconstruction and semantic prediction share a
multi-view backbone. Pixelwise class distributions are bound to the corresponding Gaussians.
Orange dashed arrows indicate training supervision.}
\label{supp:fig:sfg_architecture}
\end{figure*}

SFG initializes its reconstruction branch from a pretrained feed-forward driving-scene model~\cite{recondrive}.
The semantic decoder~\cite{ranftl2021dpt} combines shared multi-view features with RGB and ego-frame height
derived from predicted depth. It predicts a 19-class distribution for each input pixel and binds this
distribution to the corresponding Gaussian. The resulting Gaussians jointly represent geometry, appearance,
and semantics for scene rendering and editing (\cref{supp:fig:sfg_architecture}).

For editing, the predictions are grouped into road, sidewalk, terrain, structure, pole/sign, vegetation, sky,
vehicle, and vulnerable road user (VRU). This grouping keeps ground categories separate and distinguishes
editable vehicles from VRUs.

\paragraph{Drivable-area extraction.}
We project Gaussian centers associated with ground classes from multiple views and times into a common
bird's-eye-view raster. Cells with a majority of road predictions form the initial mask. Recorded ego and
vehicle footprints supplement missing coverage. We fill enclosed gaps using sidewalk and terrain predictions
as constraints, then retain regions connected to the recorded ego path. SAIG uses the resulting drivable area
$\Omega$ for both initialization and ego-response evaluation.

\subsection{Joint Training}
\label{supp:sec:sfg_training}

We jointly fine-tune the shared backbone, depth branch, Gaussian reconstruction branch, and semantic decoder.
The segmentation teacher~\cite{borse2021inverseform} and perceptual-loss network remain fixed. Each context
window contains six-camera images at two keyframes separated by 0.5\,s, with an image resolution of
$518\times280$.

\paragraph{Semantic supervision.}
We project official nuScenes point-cloud semantic annotations into the input camera images. The source-view
loss $\mathcal{L}_{\mathrm{src}}$ supervises the semantic decoder's predictions using these projected labels.
For each annotated pixel $u$, $\mathcal{S}_u$ contains the single class specified by a fine label or all classes
compatible with a coarse label. The loss is
\begin{equation}
\mathcal{L}_{\mathrm{src}}=-\frac{1}{|\mathcal{U}_{\mathrm{src}}|}
\sum_{u\in\mathcal{U}_{\mathrm{src}}}
\log\!\left(\sum_{c\in\mathcal{S}_u}p_{u,c}\right),
\label{supp:eq:source_semantics}
\end{equation}
where $\mathcal{U}_{\mathrm{src}}$ contains annotated pixels and $p_{u,c}$ is the predicted probability of class
$c$ at pixel $u$. For coarse labels, the loss supervises the summed probability of all compatible classes.
Where projected annotations are absent, the auxiliary loss $\mathcal{L}_{\mathrm{aux}}$ uses confident
predictions from the fixed segmentation teacher. At inference, semantic predictions are produced by SFG's
decoder without the teacher model.

For the rendered-view loss $\mathcal{L}_{\mathrm{rend}}$, the target viewpoints are the six input camera views
at the first context keyframe $t_0$. For each camera, we render the Gaussians from its two input keyframes
using its pose at $t_0$. At each target pixel, the accumulated class contributions $M_{u,c}$ are normalized as
\begin{equation}
\hat p_{u,c}=\frac{M_{u,c}+\varepsilon/C}{\sum_{j=1}^{C}M_{u,j}+\varepsilon},
\qquad C=19,
\label{supp:eq:rendered_semantics}
\end{equation}
where $\varepsilon>0$ stabilizes normalization. Official nuScenes point-cloud semantic annotations projected
into these views supervise the normalized class probabilities. Gradients propagate through Gaussian geometry,
opacity, and semantic predictions. The source-view and rendered-view losses thus supervise semantic predictions
before and after Gaussian rendering, respectively. The original reconstruction objective is retained throughout
joint training.

\paragraph{Optimization.}
The pretrained branches use a smaller learning rate than the semantic decoder. Semantic gradients entering
shared features and depth-derived height are scaled by 0.1. Gradients to semantic-decoder parameters and RGB
reconstruction gradients remain unchanged. The rendered-view loss weight controls its direct gradients to
Gaussian parameters. Training uses 4 NVIDIA A800 GPUs, and \cref{supp:tab:sfg_training} summarizes the configuration.

\begin{table}[!htb]
\centering\small
\caption{SFG joint-training configuration.}
\label{supp:tab:sfg_training}
\begin{tabularx}{\columnwidth}{>{\raggedright\arraybackslash}Xr}
\toprule
Parameter & Value\\
\midrule
Context images & $2\times6$\\
Input resolution & $518\times280$\\
Predicted classes / editing groups & $19\,/\,9$\\
Optimizer & AdamW\\
Pretrained-branch learning rate & $3\times10^{-7}$\\
Semantic-decoder learning rate & $10^{-4}$\\
Weight decay & 0.01\\
Semantic-input gradient scale & 0.1\\
Auxiliary loss weight & 0.1\\
Rendered-view loss weight & 0.01\\
Global batch size & 4\\
Training / warmup steps & $4{,}000\,/\,100$\\
Learning-rate schedule & Cosine\\
\bottomrule
\end{tabularx}
\end{table}

\subsection{Effect of Joint Training}
\label{supp:sec:sfg_variants}

We compare the pretrained reconstruction model with two jointly fine-tuned models. Source-view joint
fine-tuning uses projected official nuScenes semantic labels to supervise predictions on the input images.
Full joint fine-tuning additionally supervises semantic images rendered from Gaussians. Both settings jointly
update the reconstruction model and semantic decoder while retaining the reconstruction objective. The
source-view setting serves as a jointly fine-tuned control for assessing the full training scheme. The full
setting corresponds to Ours in Table~1 of the main paper.

\Cref{supp:tab:sfg_variants} reports reconstruction quality and six-class semantic mIoU. Both fine-tuned models
are evaluated using the same projected official annotations. Semantic mIoU is averaged over road, sidewalk,
terrain, vegetation, vehicle, and VRU. Full joint fine-tuning improves input-view PSNR by 0.79\,dB over the
pretrained baseline and 0.13\,dB over the source-view control. Its semantic mIoU reaches 82.00\%, close to that
of the source-view control, while novel-view reconstruction remains close to the pretrained baseline. These
results show that SFG achieves good reconstruction quality and semantic accuracy within a unified 3D
representation, which can support safety-critical driving scenario generation.

\begin{table*}[!tb]
\centering\small\setlength{\tabcolsep}{4pt}
\caption{Effect of joint training on reconstruction and semantic quality. \ranklegend
The pretrained reconstruction model has no semantic decoder.}
\label{supp:tab:sfg_variants}
\begin{tabularx}{\textwidth}{>{\raggedright\arraybackslash}Xccccccc}
\toprule
& \multicolumn{3}{c}{Input views} & \multicolumn{3}{c}{Novel views} & Semantic (\%)\\
\cmidrule(lr){2-4}\cmidrule(lr){5-7}
Training setting & PSNR$\uparrow$ & SSIM$\uparrow$ & LPIPS$\downarrow$
& PSNR$\uparrow$ & SSIM$\uparrow$ & LPIPS$\downarrow$ & mIoU (6 classes)$\uparrow$\\
\midrule
Pretrained reconstruction & 33.26 & 0.9570 & 0.0636 & \bestnum{23.15} & \secondnum{0.6968} & \secondnum{0.2925} & \notapplicable\\
Source-view joint fine-tuning & \secondnum{33.92} & \secondnum{0.9595} & \secondnum{0.0602} & 23.11 & \bestnum{0.6969} & \bestnum{0.2923} & \bestnum{82.41}\\
Full joint fine-tuning & \bestnum{34.05} & \bestnum{0.9602} & \bestnum{0.0593} & \secondnum{23.12} & 0.6963 & 0.2927 & \secondnum{82.00}\\
\bottomrule
\end{tabularx}
\end{table*}

\subsection{Reconstruction Evaluation}
\label{supp:sec:sfg_reconstruction}

\begin{figure*}[!tb]
\centering
\includegraphics[width=\textwidth]{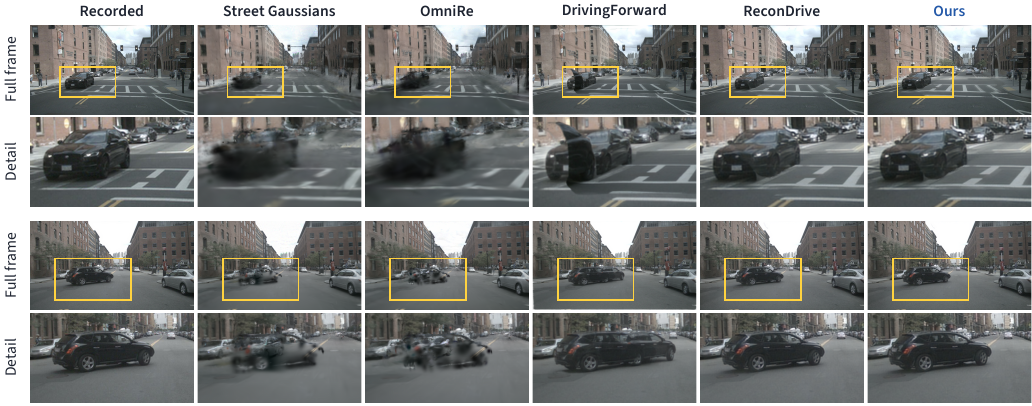}
\caption{\textbf{Input-view reconstruction of road users and road surface in nuScenes scene-0103.}}
\label{supp:fig:sfg_reconstruction}
\end{figure*}

For the quantitative results in Table~1 of the main paper, we follow the protocol in~\cite{recondrive} and report PSNR,
SSIM~\cite{wang2004ssim}, and LPIPS~\cite{zhang2018lpips} on input and held-out novel views.
\Cref{supp:fig:sfg_reconstruction} compares reconstructions at the recorded camera poses. In this qualitative
comparison, feed-forward methods use two context frames per seven-frame window. Optimization-based methods~\cite{yan2024street,chen2025omnire}
are fitted to the full sequence using their official implementations and default settings.

\subsection{Semantic Evaluation}
\label{supp:sec:sfg_semantics}

For the semantic comparison in Table~3 of the main paper, we evaluate 150 windows from 50 nuScenes validation
scenes using common reconstructed geometry. Reference labels are constructed from LiDAR, maps, and 3D boxes.
During projection, we retain the nearest visible LiDAR point per pixel. Predictions are mapped to the
evaluation categories before computing road and vehicle IoU and labeled-point accuracy at annotated locations.
For drivable-area evaluation, we project each method's ground-category predictions into a common
bird's-eye-view raster using the reconstructed geometry. Drivable-area IoU is measured against a map reference
within 30\,m.

Cross-camera agreement (Cons.) measures the fraction of corresponding pixels assigned to the same evaluation
category across cameras at the same time. We establish pixel correspondences using reconstructed depth and
camera calibration, rejecting matches that disagree with the depth in the target camera. InverseForm+vote
combines InverseForm predictions at geometrically corresponding pixels within each window using majority voting.

\begin{figure*}[!tb]
\centering
\includegraphics[width=\textwidth]{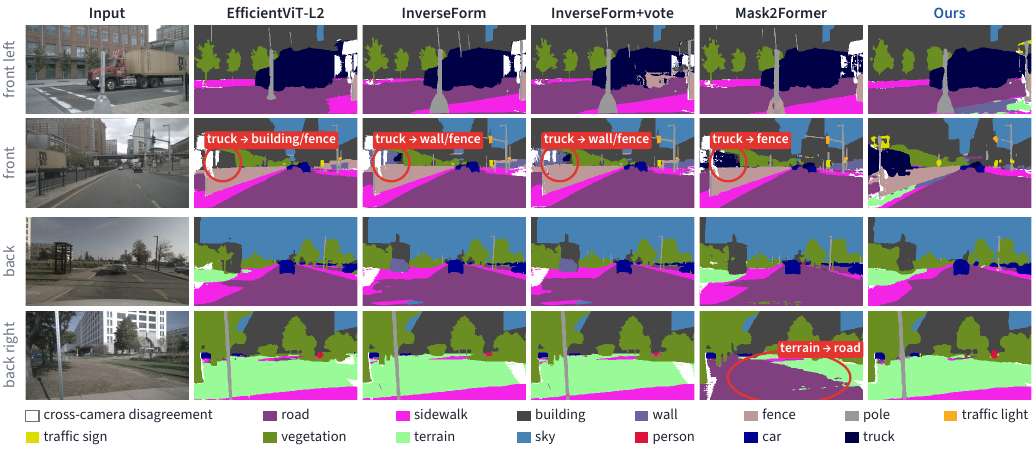}
\caption{\textbf{Cross-camera semantic predictions on nuScenes.} Each pair of rows shows neighboring cameras at the
same time. White pixels indicate inconsistent predictions for corresponding surfaces. Red ellipses mark
misclassified regions, labeled as true class $\rightarrow$ predicted class.}
\label{supp:fig:sfg_semantics}
\end{figure*}

Runtime is measured per 12-view window. For SFG, it covers the semantic decoder with shared reconstruction
features already available. \Cref{supp:fig:sfg_semantics} compares semantic predictions across neighboring
cameras in two nuScenes validation scenes.

\subsection{Zero-Shot Cross-Dataset Evaluation}
\label{supp:sec:sfg_zeroshot}

To assess SFG's cross-dataset generalization, we apply the model trained on nuScenes directly to PandaSet~\cite{xiao2021pandaset}
without fine-tuning. Each context window contains six-camera observations at two keyframes separated by
0.5\,s, matching the training setting.

\paragraph{Reconstruction.}
We evaluate all 39 sequences in the PandaSet reconstruction set using the jointly fine-tuned checkpoint
reported in Table~1 of the main paper. Frame $t$ is evaluated as the input view, and frames $t+1$ to $t+4$
as novel views, yielding 3,510 input-view and 14,040 novel-view images. All methods are evaluated against the
same $518\times280$ ground-truth images.

\begin{table}[!htb]
\centering\small\setlength{\tabcolsep}{1.2pt}
\caption{Zero-shot reconstruction on PandaSet. \ranklegend}
\label{supp:tab:zeroshot_recon}
\begin{tabularx}{\columnwidth}{>{\raggedright\arraybackslash}Xcccccc}
\toprule
& \multicolumn{3}{c}{Recon.} & \multicolumn{3}{c}{Novel}\\
\cmidrule(lr){2-4}\cmidrule(lr){5-7}
Method & {\footnotesize PSNR$\uparrow$} & {\footnotesize SSIM$\uparrow$} & {\footnotesize LPIPS$\downarrow$} & {\footnotesize PSNR$\uparrow$} & {\footnotesize SSIM$\uparrow$} & {\footnotesize LPIPS$\downarrow$}\\
\midrule
DrivingForward~\cite{tian2025drivingforward} & 24.17 & 0.817 & 0.202 & 17.15 & 0.536 & 0.410\\
ReconDrive~\cite{recondrive} & \secondnum{27.13} & \secondnum{0.883} & \bestnum{0.141} & \secondnum{20.67} & \secondnum{0.556} & \secondnum{0.392}\\
\textbf{Ours} & \bestnum{27.37} & \bestnum{0.884} & \secondnum{0.142} & \bestnum{20.72} & \bestnum{0.557} & \bestnum{0.390}\\
\bottomrule
\end{tabularx}
\end{table}

\begin{figure*}[!tb]
\centering
\includegraphics[width=\textwidth]{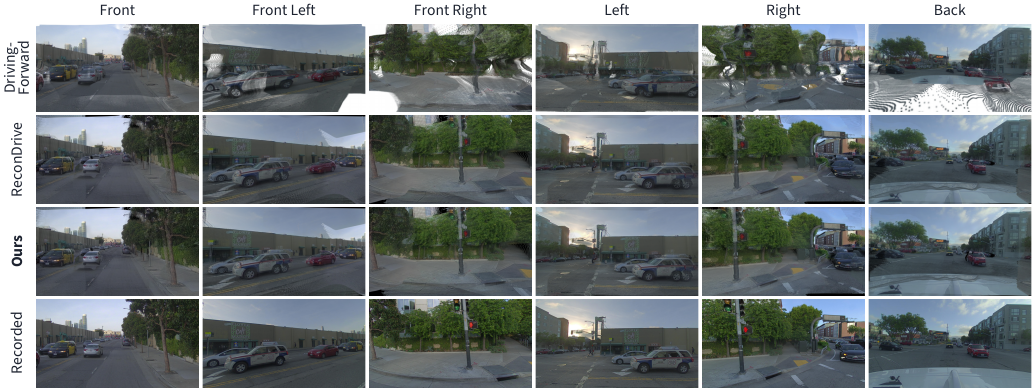}
\caption{\textbf{Zero-shot novel-view reconstruction on PandaSet.} Columns show the six cameras at one held-out time
step. The models are applied without
fine-tuning on PandaSet.}
\label{supp:fig:sfg_zeroshot}
\end{figure*}

SFG achieves the highest input-view and novel-view PSNR among the evaluated methods, at 27.37 and
20.72\,dB, respectively (\cref{supp:tab:zeroshot_recon}). Its input-view PSNR exceeds \mbox{ReconDrive} and
DrivingForward by 0.24 and 3.20\,dB, respectively. \mbox{\Cref{supp:fig:sfg_zeroshot}} compares novel-view reconstructions from \mbox{different} methods
across all six cameras.

\begin{figure*}[!tb]
\centering
\includegraphics[width=\textwidth]{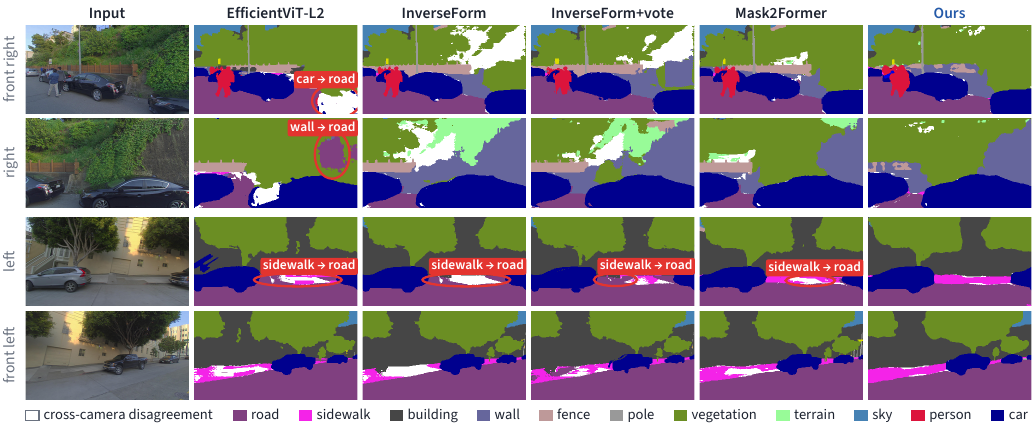}
\caption{\textbf{Zero-shot semantic predictions on PandaSet.} Each pair of rows shows neighboring cameras at the
same time. White pixels indicate inconsistent predictions for corresponding surfaces. Red ellipses mark
misclassified regions, labeled as true class $\rightarrow$ predicted class.}
\label{supp:fig:sfg_zeroshot_semantics}
\end{figure*}

\paragraph{Semantic segmentation.}
We evaluate 30 sequences with LiDAR point labels, using three windows per sequence for a total of 90 windows.
Each window contains 12 views. PandaSet labels are mapped to the semantic evaluation categories used in the
main paper and projected into images using the same visibility rule. This evaluation uses the jointly
fine-tuned SFG checkpoint.

\begin{table}[!b]
\centering\small\setlength{\tabcolsep}{2.9pt}
\caption{Zero-shot semantic segmentation on PandaSet. Road and Veh.\ denote class IoU; Acc.\ and Cons.\ denote
labeled-point accuracy and cross-camera agreement. Time is measured in ms per 12-view window. \ranklegend}
\label{supp:tab:zeroshot}
\begin{tabularx}{\columnwidth}{>{\raggedright\arraybackslash}Xccccc}
\toprule
Source & Road$\uparrow$ & Veh.$\uparrow$ & Acc.$\uparrow$ & Cons.$\uparrow$ & Time$\downarrow$\\
\midrule
EfficientViT-L2~\cite{cai2023efficientvit} & 0.751 & 0.892 & 0.838 & 0.920 & 382.5\\
InverseForm~\cite{borse2021inverseform} & \secondnum{0.856} & \secondnum{0.940} & \secondnum{0.889} & 0.943 & \secondnum{193.6}\\
InverseForm+vote & 0.837 & 0.923 & 0.875 & \bestnum{0.984} & 195.7\\
Mask2Former~\cite{cheng2022mask2former} & 0.838 & 0.937 & 0.880 & 0.943 & 819.9\\
\textbf{Ours} & \bestnum{0.886} & \bestnum{0.942} & \bestnum{0.906} & \secondnum{0.958} & \bestnum{62.3}\\
\bottomrule
\end{tabularx}
\end{table}

{\widowpenalty=10000\looseness=-1
SFG achieves the highest road IoU, vehicle IoU, and labeled-point accuracy, with cross-camera agreement of
95.8\% (\cref{supp:tab:zeroshot}). With shared reconstruction features available, its semantic decoder takes
62.3\,ms per window, \mbox{corresponding} to a $3.1$--$13.2\times$ speedup over the standalone segmentation baselines.
\Cref{supp:fig:sfg_zeroshot_semantics} compares zero-shot semantic predictions across neighboring cameras in two
PandaSet sequences. Together, the reconstruction and semantic results demonstrate SFG's cross-dataset generalization on
PandaSet.\par}

\section{Scene-Adaptive Interaction Generation}
\label{supp:sec:saig}

SAIG uses the drivable area $\Omega$ from \cref{supp:sec:sfg_architecture} and recorded surrounding traffic.
This section details scene-adaptive initialization, trajectory search, and ego-response evaluation. We then
examine initialization and full-scene solvability guidance through ablations and compare SFG-derived and
map-derived drivable areas.

\subsection{Initialization and Trajectory Search}
\label{supp:sec:saig_details}

\paragraph{Scene-adaptive initialization.}
A candidate state $\xi=(t_s,\mathbf{x},\psi,v)$ specifies the adversarial vehicle's launch time, position,
heading, and speed. Eligible launch times are recorded keyframes with at least 1.5\,s of preceding history
and 1.6\,s of remaining scene data. Positions are sampled within the drivable area $\Omega$. Headings are
sampled in $10^\circ$ steps around the local road axis, and speeds are obtained by adding $-2$, $0$, or
$+2$\,m/s to a reference speed estimated from nearby traffic.

Candidates that pass road-coverage and traffic-clearance checks are ranked using the initialization score
defined in the main paper,
\begin{equation}
s_{\mathrm{init}}(\xi)=\mathbf{w}^{\top}\boldsymbol{\phi}(\xi;\Omega,\mathcal{B}),
\label{supp:eq:initialization_score}
\end{equation}
where $\mathcal{B}$ denotes surrounding traffic following recorded trajectories. \Cref{supp:tab:saig_weights}
lists the features in $\boldsymbol{\phi}$ and their fixed weights $\mathbf{w}$. We remove near-duplicate
states while retaining candidates across launch times and travel directions. At most two candidates are
retained per launch time, with opposite-direction candidates limited to 25\%.

\paragraph{Trajectory search.}
Starting from the selected states, a pretrained trajectory diffusion prior~\cite{challenger} proposes candidate
adversarial trajectories. At each replanning step, candidates are ranked using
$J(\tau)=J_{\mathrm{base}}(\tau)+\lambda_{\mathrm{resp}}s_{\mathrm{resp}}(\tau)$, as defined in the main paper.
The base score penalizes road violations and collisions with surrounding traffic, while the response score
favors conflicts with feasible ego responses. During refinement, lower-scoring trajectories are more likely
to be resampled. The resampled trajectories are perturbed with noise, denoised again by the diffusion prior,
and re-evaluated before the next refinement round. After refinement, the adversarial vehicle follows the
lowest-scoring trajectory until the next replanning step. \Cref{supp:tab:saig_parameters} summarizes the
initialization and search settings.

\begin{table}[t]
\centering\small\setlength{\tabcolsep}{4pt}
\caption{SAIG initialization, search, and ego-response settings.}
\label{supp:tab:saig_parameters}
\begin{tabularx}{\columnwidth}{>{\hsize=0.9\hsize\raggedright\arraybackslash}X>{\hsize=1.1\hsize\raggedright\arraybackslash}X}
\toprule
Parameter & Value\\
\midrule
Launch times & 2\,Hz keyframes; $t_s\leq14$\,s\\
Position grid & 8--18\,m radius; 2\,m radial and $15^\circ$ bearing steps\\
Traffic clearance & 0.2\,m\\
Deduplication thresholds & 3\,m / 1.5\,m/s / $30^\circ$\\
Candidates per scene & 12\\
Trajectory samples & 256\\
Refinement iterations & 5\\
Replanning interval & 1\,s\\
Response weight $\lambda_{\mathrm{resp}}$ & 2.5\\
No-response penalty $\gamma$ & 3\\
Ego-response acceleration limit & 9.43\,m/s$^2$\\
Ego-response grid steps & Acceleration: 0.25\,m/s$^2$; direction: $15^\circ$; lateral offset: 0.5\,m\\
\bottomrule
\end{tabularx}
\end{table}

\begin{table}[t]
\centering\small\setlength{\tabcolsep}{4pt}
\caption{Initialization features and fixed weights. The $q$ terms are normalized scores.
Radius preference favors intermediate sampling radii.}
\label{supp:tab:saig_weights}
\begin{tabularx}{\columnwidth}{>{\raggedright\arraybackslash}Xcr}
\toprule
Feature & Symbol & Weight\\
\midrule
Road coverage & $c_{\mathrm{road}}$ & 4\\
Road-axis support & $s_{\mathrm{axis}}$ & 1\\
Road-axis confidence & $q_{\mathrm{axis}}$ & 0.5\\
Nearby moving-traffic support & $q_{\mathrm{traffic}}$ & 0.5\\
Sampling-radius preference & $q_{\mathrm{radius}}$ & 0.5\\
Speed consistency & $q_{\mathrm{speed}}$ & 0.5\\
Traffic-direction alignment & $\mathbb{1}_{\mathrm{aligned}}$ & 0.35\\
Road-boundary clearance & $q_{\mathrm{boundary}}$ & 0.75\\
Pedestrian-crossing overlap & $f_{\mathrm{crossing}}$ & $-0.5$\\
\bottomrule
\end{tabularx}
\end{table}

\subsection{Ego Responses and Interaction Retention}
\label{supp:sec:saig_responses}

Ego responses adjust longitudinal progress and lateral displacement relative to the recorded ego path. At
time $t$ after a response begins, the ego position is
\begin{equation}
\mathbf{x}_{\mathrm{resp}}(t)=\mathbf{r}\!\left(s(t)\right)+d(t)\mathbf{n}\!\left(s(t)\right),
\label{supp:eq:ego_response_construction}
\end{equation}
where $\mathbf{r}(s)$ is the recorded path parameterized by arc length and $\mathbf{n}(s)$ is the lateral
unit vector defined by the recorded heading. We obtain longitudinal progress $s(t)$ by adjusting the recorded
speed with acceleration or braking. Lateral motion accelerates and then decelerates to reach its target
offset, with zero lateral velocity at the start and completion. The magnitude and direction of the added
acceleration determine its longitudinal and lateral components.

Response onset times are sampled at or after adversarial launch and before the first collision with the ego
vehicle following its recorded trajectory. \Cref{supp:tab:saig_parameters} gives the grid resolutions.
Additional acceleration magnitudes extend up to 9.43\,m/s$^2$, and lateral target offsets up to 15\,m.

A feasible response must remain within $\Omega$, avoid the adversarial vehicle and surrounding traffic,
and satisfy the motion constraints. Surrounding vehicles follow their recorded trajectories. For each onset
$k$, $a^\star(k\mid\tau,\mathcal{C})$ is the minimum additional acceleration magnitude among feasible
responses and is infinite if none is found.

We retain an interaction when its adversarial trajectory satisfies road and motion constraints, avoids
collisions with surrounding traffic, and meets
\begin{equation}
c(\tau)=f(\tau)=1,
\label{supp:eq:interaction_retention}
\end{equation}
where $c$ indicates a collision with the ego vehicle following its recorded motion, and $f$ indicates that
a feasible ego response is found at a sampled onset. The interval from launch time $t_s$ to the last feasible
response onset $t_e$ determines the candidate keyframes for pseudo-expert labeling. The retained responses
provide trajectory candidates for this process, as detailed in \cref{supp:sec:pseudo_expert}.

\paragraph{Evaluation metrics.}
Valid-output and no-response rates are reported as percentages,
\begin{equation}
\begin{aligned}
R_{\mathrm{valid}}&=100\frac{N_{\mathrm{valid}}}{N_{\mathrm{attempt}}},\\
R_{\mathrm{noresp}}&=100\frac{N_{\mathrm{noresp}}}{N_{\mathrm{conflict}}}.
\end{aligned}
\label{supp:eq:saig_rates}
\end{equation}
Here, $N_{\mathrm{valid}}$ counts retained interactions, and $N_{\mathrm{attempt}}$ counts all generation
attempts, including initialization failures. $N_{\mathrm{conflict}}$ counts adversary-compliant outputs that
collide with the recorded ego motion after launch. $N_{\mathrm{noresp}}$ counts those for which no feasible
response is found within the evaluated response family.
The valid-output rate measures overall generation success, whereas the no-response rate focuses on
ego-response feasibility among the generated conflicts.

\subsection{Ablation Studies}
\label{supp:sec:saig_ablation}

We compare five SAIG configurations on 50 scenes selected from the nuScenes validation split
(\cref{supp:tab:saig_all}). All generated outputs are evaluated under full-scene constraints.

\begin{table*}[!tb]
\centering\small
\caption{SAIG ablation on 50 selected validation scenes, with three attempts per configuration and scene.
Conflict counts include only adversary-compliant outputs. No response is reported per conflict, Valid per
attempt, and $H$ in bits. \ranklegend}
\label{supp:tab:saig_all}
\begin{tabularx}{\textwidth}{c>{\raggedright\arraybackslash}Xlcccc}
\toprule
ID & Initialization & Guidance & Conflicts & No response (\%)$\downarrow$
& Valid (\%)$\uparrow$ & $H\uparrow$\\
\midrule
N & Non-adaptive legal & Full scene & 66 & \secondnum{3.0} & 42.7 & 2.01\\
A & Pool-random selection & Adversary only & 80 & 20.0 & 42.7 & 2.08\\
B & Pool-random selection & Full scene & 82 & 3.7 & 52.7 & 2.02\\
C & Adaptive selection & Adversary only & 107 & 15.0 & \secondnum{60.7} & \secondnum{2.24}\\
D & Adaptive selection & Full scene & 105 & \bestnum{1.0} & \bestnum{69.3} & \bestnum{2.26}\\
\bottomrule
\end{tabularx}
\end{table*}

\paragraph{Initialization.}
With full-scene solvability guidance fixed, we compare three initialization settings. Non-adaptive legal
searches a predefined grid for a legal initial state. Pool-random selection and adaptive selection use the same
legal candidate pool, but only adaptive selection ranks candidates using $s_{\mathrm{init}}$. Compared with
pool-random selection, adaptive initialization increases the conflict count from 82 to 105 and the valid-output
rate from 52.7\% to 69.3\%. Interaction-type entropy also rises from 2.02 to 2.26. These results show that
adaptive initialization improves the success rate of generating valid interactions and produces a more balanced
distribution of interaction types.

\paragraph{Solvability guidance.}
We hold the initialization strategy fixed and vary the constraints used to evaluate ego responses during
search. The control considers only the adversarial vehicle, while full-scene guidance also accounts for roads
and surrounding traffic. With pool-random selection, full-scene guidance reduces the no-response rate from
20.0\% to 3.7\%. With adaptive initialization, it reduces this rate from
15.0\% to 1.0\%. These results show that full-scene solvability guidance improves the solvability of generated
interactions while maintaining comparable conflict counts.

\paragraph{Interaction-type entropy.}
We compute
\begin{equation}
H=-\sum_{\ell=1}^{6}p_\ell\log_2p_\ell.
\label{supp:eq:interaction_entropy}
\end{equation}
For each valid interaction, we divide its weight equally among its assigned types. We average the resulting
type proportions within each scene, then equally across scenes with valid outputs, to obtain $p_\ell$.
Settings N, B, and D correspond to Table~4(a) of the main paper, and settings C and D correspond to Table~4(b).

\paragraph{Generation outcomes.}
\Cref{supp:tab:saig_outcomes} groups the same attempts by outcome. Checks proceed through initialization,
adversarial-trajectory compliance, conflict formation, and ego-response feasibility. An attempt is counted
at the first check it fails, or as valid if it passes all checks.

\begin{table}[!htb]
\centering\small\setlength{\tabcolsep}{4pt}
\caption{Generation outcomes for the configurations in \cref{supp:tab:saig_all}. Each configuration has
150 attempts. Categories follow the check order and are mutually exclusive.}
\label{supp:tab:saig_outcomes}
\begin{tabularx}{\columnwidth}{>{\raggedright\arraybackslash}Xrrrrr}
\toprule
Outcome & N & A & B & C & D\\
\midrule
Initialization failed & 3 & 0 & 0 & 0 & 0\\
Adversary constraints violated & 76 & 68 & 65 & 41 & 43\\
No conflict & 5 & 2 & 3 & 2 & 2\\
No feasible ego response found & 2 & 16 & 3 & 16 & 1\\
Valid interaction & 64 & 64 & 79 & 91 & 104\\
\midrule
Total attempts & 150 & 150 & 150 & 150 & 150\\
\bottomrule
\end{tabularx}
\end{table}

\subsection{Effect of the Drivable-Area Source}
\label{supp:sec:road_sources}

We derive drivable areas from SFG semantic segmentation or HD maps and compare SAIG's adversarial
interaction generation performance on the nuScenes validation split. We measure time for scene preparation
and trajectory generation, scene coverage, and interaction diversity. The SFG results are those reported in
the main paper.

\begin{table}[!htb]
\centering\small
\caption{Effect of the drivable-area source. Coverage is computed over the 150 validation scenes.
Time covers scene preparation and trajectory generation, in seconds per valid scenario.}
\label{supp:tab:road_sources}
\begin{tabularx}{\columnwidth}{>{\raggedright\arraybackslash}Xcc}
\toprule
Metric & Map & SFG\\
\midrule
Time (s)$\downarrow$ & 40.1 & 62.9\\
Scene coverage (\%)$\uparrow$ & 80.0 & 79.3\\
Interaction-type entropy $H\uparrow$ & 2.36 & 2.32\\
\bottomrule
\end{tabularx}
\end{table}

\begin{figure*}[!tb]
\centering
\includegraphics[width=\textwidth]{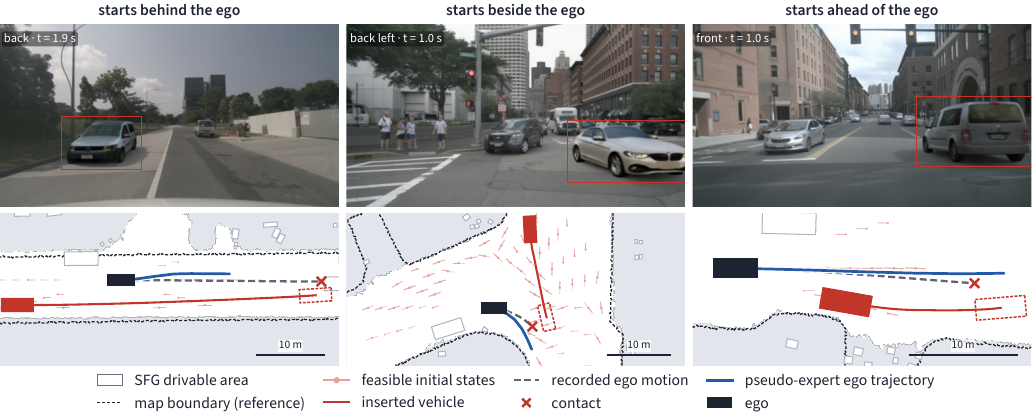}
\caption{\textbf{Scene constraints and generated interactions using SFG-derived drivable areas.}
Times are relative to launch. BEVs use the ego frame at launch, with forward pointing right. Red and gray
trajectories extend to contact, while blue pseudo-expert ego trajectories span 3\,s. Black dashed lines indicate HD-map boundaries.}
\label{supp:fig:saig_cases}
\end{figure*}

Although the SFG setting takes longer than the HD-map setting, it achieves similar scene coverage and
interaction diversity with a short runtime (\cref{supp:tab:road_sources}). These results indicate that SFG's
semantic estimates provide effective drivable-area constraints for SAIG, reducing reliance on HD-map
annotations for scenario generation. In the three cases in \cref{supp:fig:saig_cases}, initialization,
trajectory generation, ego-response evaluation, and pseudo-expert labeling all use SFG-derived drivable areas.

\section{Coordinated Generation Scheduling and Multi-View Observation Synthesis}
\label{supp:sec:cgs}

This section provides details of CGS's shared visual encoding and adaptive image harmonization. We describe
vehicle asset selection and placement, examine harmonization results, and evaluate generation time and
observation quality.

\subsection{Shared Encoding and Adaptive Harmonization}
\label{supp:sec:cgs_details}

\begin{figure*}[!t]
\centering
\includegraphics[width=\textwidth]{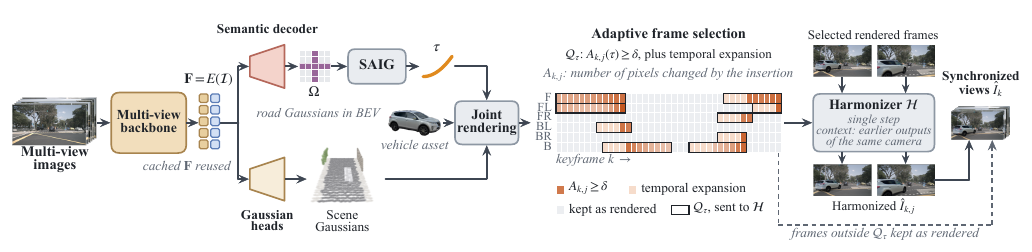}
\caption{\textbf{CGS overview.} Shared backbone features support semantic prediction and Gaussian reconstruction.
Outlined cells mark frames selected for harmonization, including those added for temporal context.
Other frames retain their rendered appearance.}
\label{supp:fig:cgs}
\end{figure*}

\paragraph{Shared visual encoding.}
Semantic prediction and Gaussian reconstruction use the same multi-view backbone. CGS retains the features
computed for semantic prediction and reuses them for subsequent Gaussian reconstruction. This avoids encoding
the same inputs twice (\cref{supp:fig:cgs}).

\paragraph{Adaptive image harmonization.}
For each trajectory $\tau$ retained by SAIG, we jointly render the reconstructed scene and the adversarial
vehicle asset. CGS measures the visible effect of the edit by comparing each rendered frame $\tilde I_{k,j}$
with its counterpart $\tilde I^0_{k,j}$ without the adversarial vehicle. With RGB values normalized to $[0,1]$,
the changed-pixel count is
\begin{equation}
A_{k,j}(\tau)=\sum_u\mathbb{1}\!\left[
\left\|\tilde I_{k,j}(u)-\tilde I^0_{k,j}(u)\right\|_\infty>\eta\right].
\label{supp:eq:changed_pixels}
\end{equation}
Here, $u$ indexes pixels and $\eta$ is the color-difference threshold. We use $\eta=3/255$ and $\delta=638$
pixels on the $518\times280$ rendering grid. Frames satisfying $A_{k,j}(\tau)\geq\delta$ are selected. We add
the four preceding frames to each selected segment to form the harmonization set $\mathcal{Q}_\tau$.

A pretrained single-step diffusion harmonizer~\cite{zhang2026diffusionharmonizer} processes the frames in
$\mathcal{Q}_\tau$ in temporal order. Previously harmonized frames from the same camera provide temporal
context. Frames outside $\mathcal{Q}_\tau$ retain their rendered appearance, and outputs from all cameras
form the synchronized multi-view observations.

\subsection{Vehicle Assets and Harmonization Results}
\label{supp:sec:harmonization}

\begin{figure*}[!tb]
\centering
\includegraphics[width=\textwidth]{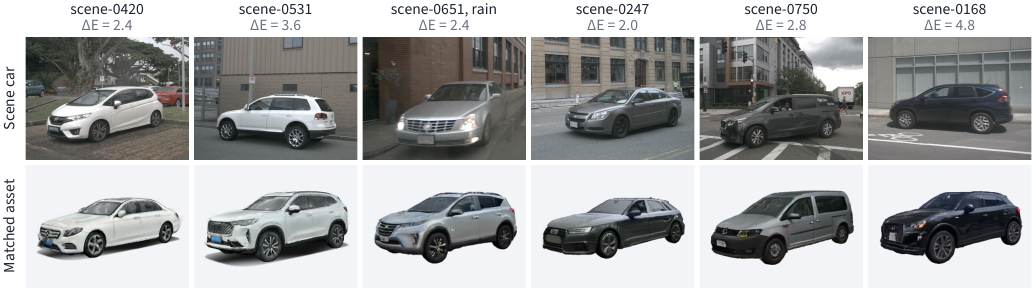}
\caption{\textbf{Scene-matched vehicle assets.} Each column shows a scene reference vehicle (top) and the matched
asset rendered in isolation (bottom). $\Delta E$ denotes the CIELAB distance between their appearance descriptors.}
\label{supp:fig:asset_matching}
\end{figure*}

\paragraph{Vehicle assets and placement.}
We use a library of ten Gaussian vehicle assets spanning multiple vehicle types~\cite{du2025realcar,cao2026assetharvester}.
Color-cast correction reduces color biases inherited from the capture environment. For scene matching,
we extract color descriptors from a reference vehicle in the recorded scene and from standalone asset
renderings. Each descriptor is the median CIELAB color of the brighter half of the vehicle pixels.
The selected asset is
\begin{equation}
m^\star=\arg\min_{m\in\mathcal{A}}\left\|\mathbf{d}_m-\mathbf{d}_{\mathrm{ref}}\right\|_2,
\label{supp:eq:asset_matching}
\end{equation}
where $\mathcal{A}$ denotes the asset library, and $\mathbf{d}_m$ and $\mathbf{d}_{\mathrm{ref}}$ are the asset
and reference-vehicle descriptors. \Cref{supp:fig:asset_matching} compares scene reference vehicles with the
selected assets. We select the asset and fix its physical dimensions before trajectory generation, ensuring
consistency across initialization, collision checks, and rendering. During rendering, the asset follows the
generated trajectory with its base aligned to the estimated road surface.

\begin{figure*}[!tb]
\centering
\includegraphics[width=\textwidth]{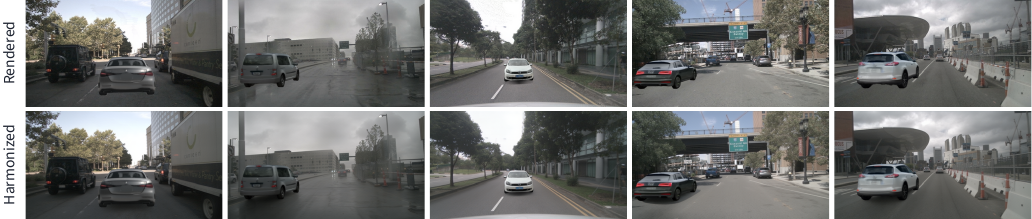}
\caption{\textbf{Image harmonization in five validation scenes.} Each column compares the same camera frame after
joint rendering (top) and harmonization (bottom).}
\label{supp:fig:harmonization}
\end{figure*}

\begin{figure*}[!tb]
\centering
\includegraphics[width=\textwidth]{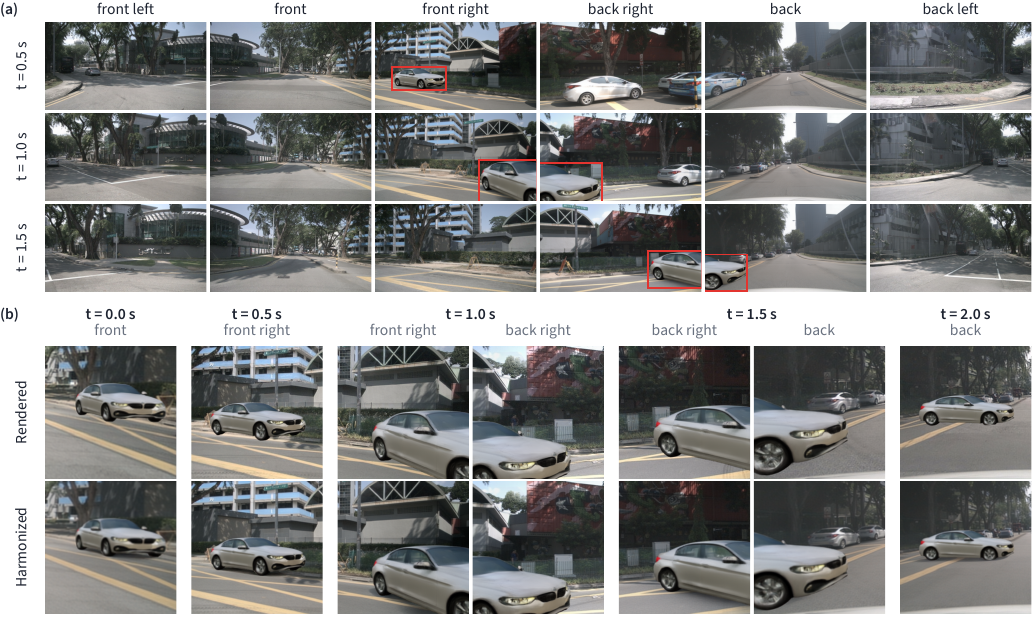}
\caption{\textbf{Multi-view harmonization over time.} The inserted vehicle approaches head-on and passes on the ego
vehicle's right, appearing in four cameras. (a) Synchronized six-camera observations at three keyframes, with the
inserted vehicle boxed. (b) Vehicle crops before (top) and after (bottom) harmonization, grouped by time.}
\label{supp:fig:harmonization_temporal}
\end{figure*}

\paragraph{Harmonization results.}
\Cref{supp:fig:harmonization} compares joint renderings and harmonized observations in five validation scenes.
After harmonization, the inserted vehicles have lighting and color more consistent with their surroundings.
\Cref{supp:fig:harmonization_temporal} follows an inserted vehicle across four cameras over time. In this
sequence, the vehicle retains a visually consistent appearance across cameras and keyframes.

\subsection{Efficiency and Quality Evaluation}
\label{supp:sec:cgs_protocol}

We compare CGS with a setting that recomputes backbone features and harmonizes all camera frames. Both settings
use identical adversarial trajectories, vehicle assets, and asset scales. Measurements are taken on a single
RTX\,4090. Time per valid scenario follows the accounting in \cref{supp:sec:generation_protocol}, excluding
model loading and pseudo-expert trajectory labeling.

Both ablation settings apply harmonization to full-frame images. For the generation-cost results in Table~2
of the main paper, RiskStage harmonizes vehicle regions and blends the resulting crops into the full frame.
Within the CGS ablation, frame selection reduces the average number of harmonized frames from 233 to 71 per run.

FID~\cite{heusel2017fid} and FVD~\cite{unterthiner2018fvd} are computed on matched six-camera outputs against
the same recorded references, using 3,726 frames and 384 video clips. We evaluate the complete observations
produced by both settings, including frames retained without harmonization.

\begin{table}[!htb]
\centering\small\setlength{\tabcolsep}{3pt}
\caption{CGS ablation and time decomposition. Time is measured in seconds per valid scenario.
Fwd.\ denotes the reconstruction forward pass after semantic prediction; Harm.\ denotes harmonization.
Other covers the remaining timed work. Red cells mark improvements over w/o CGS.}
\label{supp:tab:cgs}
\begin{tabularx}{\columnwidth}{>{\raggedright\arraybackslash}Xcccccc}
\toprule
Setting & Fwd.$\downarrow$ & Harm.$\downarrow$ & Other
& Total$\downarrow$ & FID$\downarrow$ & FVD$\downarrow$\\
\midrule
w/o CGS & 26.7 & 99.7 & 78.8 & 205.2 & 15.89 & 54.72\\
CGS & \bestnum{14.2} & \bestnum{25.1} & 83.4 & \bestnum{122.6} & \bestnum{10.65} & \bestnum{37.71}\\
\bottomrule
\end{tabularx}
\end{table}

CGS reduces generation time per valid scenario from 205.2 to 122.6\,s, a $1.67\times$ speedup
(\cref{supp:tab:cgs}). The time breakdown identifies harmonization as the largest source of savings, with a
further reduction in the reconstruction forward pass. Lower FID and FVD indicate improved observation quality,
consistent with the overall ablation reported in Table~5 of the main paper.
\section{Scenario Generation Evaluation Protocol}
\label{supp:sec:generation_protocol}

This section supplements the generation evaluation in Table~2 and Figures~4 and~5 of the main paper.

\Needspace*{5\baselineskip}
\paragraph{Generation cost and scene coverage.}
We measure generation time on a workstation with an Intel Core i9-14900K CPU and a single RTX\,4090 GPU.
The time per valid scenario (2\,Hz keyframes) is computed as
\begin{equation}
T_{\mathrm{valid}}=\frac{\sum_s T_s}{\sum_s N_s^{\mathrm{valid}}},
\label{supp:eq:valid_time}
\end{equation}
where $T_s$ is the accumulated generation time for scene $s$, and $N_s^{\mathrm{valid}}$ is its number of valid
outputs. Interactions retained under the conditions in \cref{supp:sec:saig_responses} are counted as valid
RiskStage outputs. Timing covers scene preparation, trajectory search, rendering, and image harmonization,
excluding model loading and pseudo-expert ego trajectory generation. Scene preparation is counted once per
scene, and all generation attempts contribute to the measured time. Baseline timings include trajectory
generation and multi-view video synthesis.

Scene coverage is the fraction of the 150 nuScenes validation scenes with at least one valid generated output.
Each scene is counted once, regardless of the number of generated outputs. Baseline coverage is computed
from the generated scenario data provided by each method~\cite{challenger,safemvdrive}.

\paragraph{Visual quality.}
We compare the final generated observations with official nuScenes~\cite{caesar2020nuscenes} RGB images from
\mbox{corresponding} scenes, timestamps, and cameras. Images are first standardized to $518\times280$ resolution.

For FID~\cite{heusel2017fid}, we resize each image to $299\times299$ using bilinear interpolation and extract
pooled Inception-v3 features. For FVD~\cite{unterthiner2018fvd}, we form 16-frame clips at 2\,Hz with a stride
of eight frames. Frames are resized to $224\times224$ using bilinear interpolation and normalized to $[0,1]$
before extracting I3D pre-softmax features~\cite{carreira2017i3d}. Clips are constructed separately for each
camera to preserve temporal order. For each metric, feature samples from all cameras are combined to estimate
the means and covariances of the generated and real distributions. We then compute their Fr\'echet distance.

Each baseline comparison uses validation scenes common to RiskStage and that baseline. For FID and FVD,
Table~2 of the main paper reports the larger RiskStage value from the two comparisons. The CGS ablation
uses the matched observations described in \cref{supp:sec:cgs_protocol}.

\paragraph{Motion and interaction diversity.}
These statistics are computed for moving adversarial vehicles. We estimate planar acceleration from
trajectory positions at 2\,Hz and compute its peak magnitude over moving trajectory segments. We report the
fraction of adversarial vehicles whose peak acceleration exceeds the braking-test reference of
$9.43\,\mathrm{m/s^2}$~\cite{adac2024braking}.

We classify interactions using relative position, travel direction, and lateral motion in the ego frame.
Classification is evaluated at the time of closest approach,
$t^\star=\arg\min_t\|\mathbf x_{\mathrm{adv}}(t)-\mathbf x_{\mathrm{ego}}(t)\|_2$, where
$\mathbf x_{\mathrm{adv}}$ and $\mathbf x_{\mathrm{ego}}$ are global planar positions. The six categories are
ahead, behind, alongside, cut-in, crossing, and oncoming, as in Figure~5 of the main paper. Cut-in describes
a vehicle entering the ego vehicle's travel path from the side, assessed using its lateral motion over the
preceding 2\,s. Crossing and oncoming interactions are distinguished by relative travel direction. For this
comparison, each classified trajectory contributes to one category. Interaction-type entropy is computed
from the resulting category proportions.
\FloatBarrier
\section{Generated Data and E2E Model Evaluation}
\label{supp:sec:e2e}

We describe pseudo-expert trajectory construction and the data mixtures used for E2E fine-tuning. We then
detail the evaluation settings and report open-loop, closed-loop, and cross-method results.

\subsection{Pseudo-Expert Ego Trajectories}
\label{supp:sec:pseudo_expert}

We construct pseudo-expert ego trajectories as planning labels for generated observations within the interval
$[t_s,t_e]$ determined by SAIG. At each keyframe $k$, we generate multi-view observations $\hat I_k$ at the
recorded ego pose $T_k$. The corresponding trajectory label starts from that state and covers the next 3\,s.
Let $\mathcal{K}$ denote the keyframes in this interval.

\begin{figure*}[!tb]
\centering
\includegraphics[width=\textwidth]{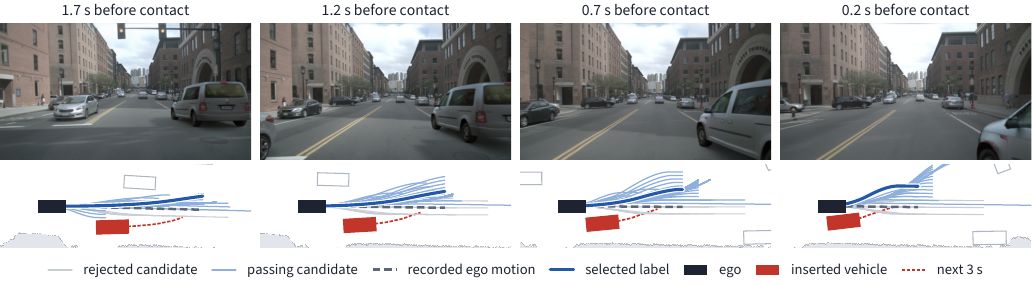}
\caption{\textbf{Pseudo-expert trajectory construction.} Four consecutive keyframes show an inserted vehicle
cutting in from the right. The top row contains harmonized front-camera observations. The bottom row compares
candidate and selected trajectories over the 3\,s planning horizon, in the ego frame with forward pointing right
and the drivable area in white. Selected trajectories deviate leftward from the recorded ego motion while
maintaining a consistent avoidance decision across keyframes.}
\label{supp:fig:pseudo_expert}
\end{figure*}

\paragraph{Candidate construction.}
At keyframe $k$, the candidate set is
\begin{equation}
\mathcal{Y}_k=\{y_k^{\mathrm{rec}}\}\cup\mathcal{Y}_k^{\mathrm{IDM}}\cup\mathcal{Y}_k^{\mathrm{SAIG}},
\label{supp:eq:pseudo_candidates}
\end{equation}
where $y_k^{\mathrm{rec}}$ is the recorded ego trajectory, $\mathcal{Y}_k^{\mathrm{IDM}}$ contains car-following
trajectories based on the Intelligent Driver Model (IDM)~\cite{dauner2023pdm,treiber2000idm}, and
$\mathcal{Y}_k^{\mathrm{SAIG}}$ contains feasible avoidance responses from SAIG. IDM candidates vary the target
speed between 20\% and 100\% of the reference speed, with lateral offsets of $-1$, $0$, and $+1$\,m. Each SAIG
avoidance decision combines a lateral target with a longitudinal action. For each decision, we retain the
feasible response requiring the least additional acceleration. All candidates start from the current ego state.

\paragraph{Constraint checks.}
Candidates are evaluated over a 3\,s planning horizon for collision avoidance, road compliance, and motion
feasibility. We check road compliance using the drivable area and the corridor swept by recorded ego motion.
Generated ego maneuvers satisfy a total acceleration bound of $9.43\,\mathrm{m/s^2}$. We also use the
time-to-collision (TTC) projection check from PDM-Closed~\cite{dauner2023pdm} to filter candidates based on
short-term motion extrapolation. If this check already fails at the initial state shared by all candidates,
it is not used to reject candidates. Collision avoidance and road compliance remain hard constraints. Beyond the planning horizon, we check whether a feasible response remains available
to avoid the inserted vehicle.

\paragraph{Joint selection.}
Let $\mathcal{F}_k\subseteq\mathcal{Y}_k$ denote the candidates that pass the checks. We jointly select
$Y=(y_k)_{k\in\mathcal{K}}$ by
\begin{equation}
\begin{aligned}
Y^\star&=\underset{\substack{y_k\in\mathcal{F}_k\cup\{\varnothing\}\\k\in\mathcal{K}}}
{\operatorname{arg\,min}^{\mathrm{lex}}}\\
&\quad\bigl(N_{\varnothing}(Y),\,N_{\mathrm{switch}}(Y),\,-D(Y),\,E(Y)\bigr).
\end{aligned}
\label{supp:eq:pseudo_joint_selection}
\end{equation}
{\looseness=-1
Here, $\varnothing$ denotes an unlabeled keyframe, and $\mathrm{lex}$ specifies optimization in the listed
priority order. We first minimize the number of unlabeled keyframes $N_{\varnothing}$, then the number of
changes in avoidance decision $N_{\mathrm{switch}}$. We next maximize $D$, the sum of minimum clearances along
the selected trajectories, each capped at 0.48\,m. Finally, we minimize $E$, the sum of acceleration-based
maneuver costs.\par}

The supervision window $\mathcal{W}=\{k\in\mathcal{K}:y_k\neq\varnothing\}$ contains keyframes with
pseudo-expert trajectory labels within $[t_s,t_e]$. Within this interval, only these labeled keyframes contribute
to the planning loss. Outside $\mathcal{K}$, recorded ego trajectories are retained as training labels when
they pass the same checks. \Cref{supp:fig:pseudo_expert} compares the candidates and selected trajectories across
keyframes.

\subsection{Dataset Composition and Fine-Tuning}
\label{supp:sec:finetuning}

We construct original and adversarial datasets following the official nuScenes split~\cite{caesar2020nuscenes}.
The original and adversarial test sets are built from its 150 validation scenes. The adversarial training and
test sets contain 549 and 119 adversarial scenarios, respectively. We use original-scene renderings for the
remaining scenarios to match the official split sizes.

Let $P_{\mathrm{orig}}$ and $P_{\mathrm{adv}}$ denote the training sampling distributions over the original and
adversarial dataset versions, respectively. Fine-tuning with $\rho\in\{0,0.25,0.5,0.75,1\}$ uses
\begin{equation}
P_\rho=(1-\rho)P_{\mathrm{orig}}+\rho P_{\mathrm{adv}}.
\label{supp:eq:mixture_distribution}
\end{equation}
The mixture proportion specifies the fraction of training samples drawn from the adversarial dataset version.
Sampling is balanced across scenes within each dataset version. The $\rho=0$ control uses only original-scene
renderings, while $\rho=1$ samples exclusively from the adversarial version, including the original-scene
renderings used to complete the split. Evaluation sets remain fixed across mixtures.

We fine-tune UniAD~\cite{hu2023uniad}, VAD~\cite{jiang2023vad}, and DiffusionDrive~\cite{liao2025diffusiondrive}
from their official pretrained weights. All models use AdamW with linear warmup followed by cosine decay.
Trainable image-backbone parameters use $0.1\times$ the base learning rate.

{\looseness=-1
Every run processes 160,032 training targets, with the number of updates determined by the global batch size.
For each model and adversarial-data proportion, we evaluate the final checkpoint after completing this
training budget. Training uses 8 NVIDIA A800 GPUs, a 3\,s planning horizon, and random seed 42.
\Cref{supp:tab:e2e_settings} summarizes the model-specific settings. Batch denotes the global batch size;
Updates and Warmup denote total and warmup optimizer steps.\par}

UniAD freezes its image backbone, neck, BEV encoder, and batch-normalization (BN) layers. VAD freezes the
backbone stem, first residual stage, and BN layers. DiffusionDrive keeps the backbone and BN layers trainable.

\begin{table}[!htb]
\centering\small\setlength{\tabcolsep}{4pt}
\caption{E2E fine-tuning settings. LR denotes the base learning rate, and WD denotes weight decay.}
\label{supp:tab:e2e_settings}
\begin{tabular*}{\columnwidth}{@{}l@{\extracolsep{\fill}}ccccc@{}}
\toprule
Model & LR & WD & Batch & Updates & Warmup\\
\midrule
UniAD & $2\times10^{-5}$ & 0.01 & 8 & 20,004 & 500\\
VAD & $2\times10^{-5}$ & 0.01 & 8 & 20,004 & 500\\
DiffusionDrive & $3\times10^{-5}$ & 0.001 & 48 & 3,334 & 83\\
\bottomrule
\end{tabular*}
\end{table}

\subsection{Open-Loop and Closed-Loop Evaluation}
\label{supp:sec:e2e_protocol}

\paragraph{Open-loop evaluation.}
{\looseness=-1
We evaluate the models on the original and adversarial test sets. At each valid keyframe with 3\,s of future
data, the predicted ego trajectory is checked at 10\,Hz. A collision is recorded if the predicted ego box
intersects another road user's box at any sampled instant. Collision rates are averaged over valid planning
starts within each scene and then equally across scenes. All checkpoints use the same input observations and
collision detection method for each test set.\par}

\begin{table}[!t]
\centering\small\setlength{\tabcolsep}{3pt}
\caption{Open-loop results across data mixtures on the adversarial and original test sets.
Bold values denote nonzero mixtures with lower values than the $0\%$ control for the same model and metric.
Official denotes the pretrained checkpoint.}
\label{supp:tab:e2e_mixtures}
\begin{tabularx}{\columnwidth}{>{\raggedright\arraybackslash}X*{6}{r}}
\toprule
Model & Official & 0\% & 25\% & 50\% & 75\% & 100\%\\
\midrule
\multicolumn{7}{l}{\textit{Adversarial, collision rate (\%)$\downarrow$}}\\
UniAD & 19.51 & 19.28 & \textbf{17.62} & \textbf{18.06} & \textbf{17.01} & \textbf{16.94}\\
VAD & 20.44 & 21.37 & \textbf{20.02} & \textbf{20.02} & \textbf{19.84} & \textbf{19.53}\\
DiffusionDrive & 18.77 & 19.08 & \textbf{17.87} & \textbf{17.75} & \textbf{17.18} & \textbf{16.71}\\
\midrule
\multicolumn{7}{l}{\textit{Original, collision rate (\%)$\downarrow$}}\\
UniAD & 2.46 & 2.21 & \textbf{2.17} & 2.25 & \textbf{2.09} & \textbf{2.15}\\
VAD & 2.05 & 2.05 & \textbf{1.97} & \textbf{1.99} & \textbf{1.98} & \textbf{1.76}\\
DiffusionDrive & 1.86 & 1.72 & \textbf{1.39} & \textbf{1.52} & \textbf{1.29} & \textbf{1.29}\\
\midrule
\multicolumn{7}{l}{\textit{Original, 3\,s FDE (m)$\downarrow$}}\\
UniAD & 2.12 & 1.92 & 1.92 & 1.92 & \textbf{1.91} & 1.98\\
VAD & 2.04 & 1.93 & \textbf{1.90} & 1.98 & 2.00 & 1.99\\
DiffusionDrive & 1.87 & 1.96 & \textbf{1.87} & \textbf{1.93} & \textbf{1.94} & \textbf{1.93}\\
\bottomrule
\end{tabularx}
\end{table}

\paragraph{Closed-loop evaluation.}
We evaluate UniAD, VAD, and DiffusionDrive on reconstructions of ten original nuScenes scenarios using an
adapted and refined HUGSIM closed-loop simulator~\cite{zhou2024hugsim}. We test official pretrained checkpoints
and checkpoints fine-tuned with different adversarial-data proportions. The ego follows recorded motion for 1.5\,s before the model takes control at
2\,Hz using six-camera renderings.

\begin{figure*}[!tp]
\centering
\includegraphics[width=0.92\textwidth]{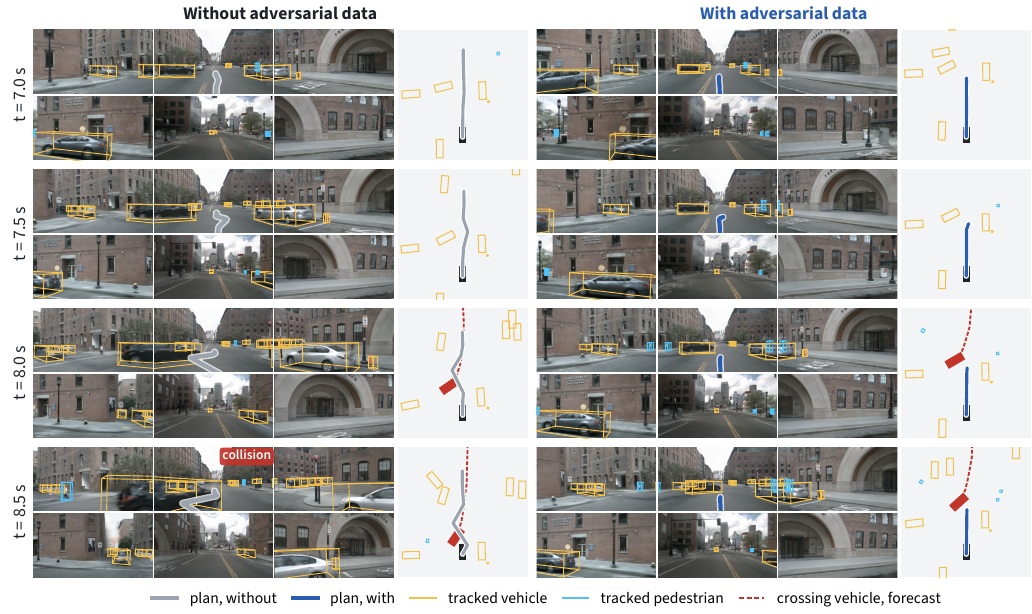}
\caption{\textbf{Closed-loop planning comparison.} UniAD with the 0\% control (left) and 50\% RiskStage mixture
(right), shown at matched simulation times. Each row includes six-camera observations and a bird's-eye view of
tracked objects, the crossing vehicle's predicted motion, and the 3\,s plan. Both checkpoints detect the crossing
vehicle. The control continues and collides at 8.5\,s, whereas the fine-tuned checkpoint shortens its plan and yields.}
\label{supp:fig:closedloop_case}
\par\vspace{10pt}
\begin{minipage}{\textwidth}
\centering\small\renewcommand{\arraystretch}{1.2}
\captionof{table}{Closed-loop scores on ten HUGSIM scenarios. All metrics are higher-is-better. Bold values denote nonzero mixtures that exceed the $0\%$ control for the same model and metric. Official denotes the pretrained checkpoint.}
\label{supp:tab:closedloop}
\begin{tabular*}{\textwidth}{@{\extracolsep{\fill}}llccccccc}
\toprule
Model & Weights & HD-Score & RC & PDMS & NC & DAC & TTC & C\\
\midrule
\multirow{6}{*}{UniAD} & Official & 0.288 & 0.388 & 0.659 & 0.735 & 1.000 & 0.786 & 0.833\\
 & 0\% & 0.331 & 0.459 & 0.649 & 0.749 & 1.000 & 0.804 & 0.806\\
 & 25\% & \textbf{0.413} & \textbf{0.575} & 0.611 & \textbf{0.771} & 0.980 & \textbf{0.818} & 0.684\\
 & 50\% & \textbf{0.377} & \textbf{0.478} & \textbf{0.661} & \textbf{0.778} & 0.992 & \textbf{0.853} & 0.703\\
 & 75\% & 0.238 & 0.459 & 0.446 & 0.705 & 0.878 & 0.644 & 0.404\\
 & 100\% & 0.304 & \textbf{0.541} & 0.445 & 0.574 & 0.932 & 0.644 & 0.512\\
\midrule
\multirow{6}{*}{VAD} & Official & 0.214 & 0.432 & 0.438 & 0.544 & 0.926 & 0.485 & 1.000\\
 & 0\% & 0.185 & 0.440 & 0.381 & 0.547 & 0.839 & 0.492 & 1.000\\
 & 25\% & \textbf{0.221} & \textbf{0.449} & \textbf{0.426} & \textbf{0.549} & \textbf{0.927} & 0.470 & 1.000\\
 & 50\% & \textbf{0.245} & \textbf{0.464} & \textbf{0.444} & 0.514 & \textbf{0.934} & 0.446 & 1.000\\
 & 75\% & \textbf{0.220} & \textbf{0.449} & \textbf{0.424} & 0.521 & \textbf{0.910} & 0.451 & 1.000\\
 & 100\% & \textbf{0.214} & \textbf{0.442} & \textbf{0.414} & 0.506 & \textbf{0.922} & 0.413 & 1.000\\
\midrule
\multirow{6}{*}{DiffusionDrive} & Official & 0.278 & 0.377 & 0.599 & 0.624 & 1.000 & 0.675 & 1.000\\
 & 0\% & 0.297 & 0.379 & 0.629 & 0.653 & 1.000 & 0.669 & 1.000\\
 & 25\% & \textbf{0.306} & 0.374 & \textbf{0.686} & \textbf{0.724} & 1.000 & \textbf{0.696} & 1.000\\
 & 50\% & \textbf{0.348} & \textbf{0.416} & \textbf{0.680} & \textbf{0.699} & 1.000 & \textbf{0.707} & 1.000\\
 & 75\% & 0.295 & \textbf{0.385} & 0.602 & 0.627 & 1.000 & \textbf{0.682} & 0.994\\
 & 100\% & 0.295 & \textbf{0.388} & 0.610 & 0.632 & 1.000 & \textbf{0.680} & 0.977\\
\bottomrule
\end{tabular*}
\end{minipage}
\end{figure*}

HD-Score is computed as
\begin{equation}
\mathrm{HD\text{-}Score}=\frac{1}{N}\sum_{s=1}^{N}\mathrm{RC}_s\,\mathrm{PDMS}_s,\qquad N=10,
\label{supp:eq:hd_score}
\end{equation}
where $\mathrm{RC}_s$ is the maximum fraction of the reference route completed and $\mathrm{PDMS}_s$ is the
mean PDM score over the evaluated steps in scenario $s$. At each evaluation step $t$, the PDM score is
\begin{equation}
p_t=\mathrm{NC}_t\,\mathrm{DAC}_t\frac{5\,\mathrm{TTC}_t+2\,\mathrm{C}_t}{7}.
\label{supp:eq:pdm_score}
\end{equation}
No collision (NC) checks the planned trajectory for collisions with static geometry and other traffic.
Drivable-area compliance (DAC) evaluates the vehicle footprint against the drivable area. Time to collision
(TTC) checks potential collisions under short-term motion extrapolation. Comfort (C) checks limits on
acceleration, jerk, yaw rate, and yaw acceleration. NC, TTC, and C equal 1 when their checks pass and 0 otherwise.
NC, DAC, TTC, and C are first averaged within each scenario. \Cref{supp:tab:closedloop} reports averages across
the ten scenarios.

\Cref{supp:fig:closedloop_case} compares the two UniAD checkpoints in a crossing interaction. Both detect the
crossing vehicle, but the checkpoint fine-tuned with RiskStage data yields instead of continuing into a collision.

\subsection{Results Across Fine-Tuning Mixtures}
\label{supp:sec:checkpoints}

\Cref{supp:tab:e2e_mixtures} reports open-loop results for all five mixtures, and \cref{supp:tab:closedloop}
reports closed-loop scores. For each model and mixture, the same fine-tuned checkpoint is evaluated on the
original and adversarial test sets.

At 100\%, adversarial-test collision rates decrease by 1.84--2.38 percentage points relative to the 0\% controls.
At this mixture, original-test collision rates also decrease for all three models. Among the tested mixtures,
UniAD achieves its highest closed-loop HD-Score at 25\%, while VAD and DiffusionDrive achieve theirs at 50\%.

\begin{figure*}[!t]
\centering
\includegraphics[width=0.95\textwidth]{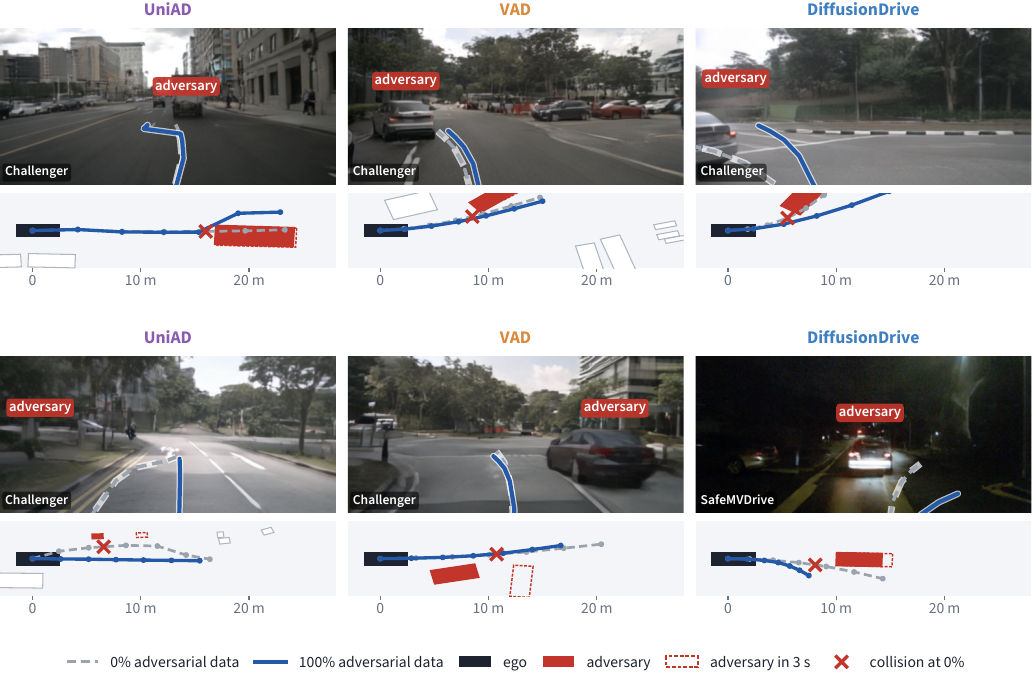}
\caption{\textbf{Open-loop plans on scenarios from comparison methods.} Each panel compares the same model's
0\% control and 100\% checkpoint using a front-camera observation and a bird's-eye view of the 3\,s plans.
Tags identify the generation method, and crosses mark the first contact of the 0\% plan.}
\label{supp:fig:cross_method_cases}
\par\vspace{10pt}
\begin{minipage}{\textwidth}
\centering\small\setlength{\tabcolsep}{4pt}
\captionof{table}{Three-second open-loop collision rates (\%) on scenarios from the comparison methods.
Bold values denote nonzero mixtures with lower collision rates than the $0\%$ control for the same model and test set.
Official denotes the pretrained checkpoint.}
\label{supp:tab:cross_method}
\begin{tabular*}{\textwidth}{@{\extracolsep{\fill}}l*{12}{r}}
\toprule
& \multicolumn{6}{c}{Challenger} & \multicolumn{6}{c}{SafeMVDrive}\\
\cmidrule(lr){2-7}\cmidrule(lr){8-13}
Model & Official & 0\% & 25\% & 50\% & 75\% & 100\% & Official & 0\% & 25\% & 50\% & 75\% & 100\%\\
\midrule
UniAD & 13.82 & 12.68 & \textbf{11.79} & \textbf{11.50} & \textbf{12.03} & \textbf{11.31}
 & 23.07 & 21.33 & \textbf{20.18} & \textbf{20.71} & \textbf{19.29} & \textbf{21.25}\\
VAD & 20.19 & 20.11 & \textbf{19.89} & \textbf{19.63} & \textbf{19.81} & \textbf{19.77}
 & 21.18 & 23.59 & \textbf{22.67} & \textbf{21.07} & \textbf{22.08} & \textbf{22.48}\\
DiffusionDrive & 4.18 & 5.03 & \textbf{3.71} & \textbf{4.28} & \textbf{4.23} & \textbf{4.12}
 & 44.79 & 44.18 & 45.25 & \textbf{43.00} & \textbf{40.91} & \textbf{40.23}\\
\bottomrule
\end{tabular*}
\end{minipage}
\end{figure*}

\subsection{Testing on Comparison-Method Scenarios}
\label{supp:sec:cross_method}

We further evaluate the models on scenarios from Challenger~\cite{challenger} and SafeMVDrive~\cite{safemvdrive}
to examine whether training benefits extend to other generated scenarios. For each model and mixture,
the same final checkpoint is evaluated on both datasets.

Collisions are detected using oriented boxes at 10\,Hz over a 3\,s horizon. Results are averaged equally across
original scenes, with SafeMVDrive results first averaged across its generated scenarios within each scene.
The evaluation covers 83 Challenger scenes and 31 SafeMVDrive scenes.

\begin{samepage}
Across the evaluated mixtures, fine-tuning with RiskStage data generally lowers collision rates for all three
E2E models on the Challenger and SafeMVDrive datasets (\cref{supp:tab:cross_method}). These results demonstrate
that the planning safety gains from training E2E models with RiskStage scenarios generalize to other
safety-critical scenario datasets.

\Cref{supp:fig:cross_method_cases} compares changes in planned paths and travel distances after fine-tuning.
\end{samepage}

 \end{document}